\pdfoutput=1
\documentclass[11pt]{article}

\usepackage{acl}

\usepackage{times}
\usepackage{latexsym}
\usepackage{multirow}
\usepackage[T1]{fontenc}
\usepackage[utf8]{inputenc}

\usepackage{microtype}

\usepackage{inconsolata}

\usepackage{graphicx}
\usepackage{booktabs}
\usepackage{fontawesome}
\usepackage{array}
\usepackage{amsmath} 
\usepackage{pifont}
\usepackage{bm}
\usepackage{xspace}
\usepackage{mathrsfs,algorithmic}
\usepackage{tcolorbox}
\usepackage{caption}
\usepackage{subcaption}
\usepackage[ruled,vlined]{algorithm2e}
\usepackage{fancyhdr}
\usepackage{marvosym}
\definecolor{darkorange}{RGB}{255, 140, 0}
\definecolor{lightgreen}{RGB}{145, 204, 117}
\definecolor{lightyellow}{RGB}{250, 200, 88}
\definecolor{lightred}{RGB}{238, 102, 102}
\definecolor{lightblue}{RGB}{115, 192, 222}
\newtcolorbox{promptbox}[2][Prompt]{
colback=black!5!white,
arc=5pt, 
boxrule=0.5pt,
fonttitle=\bfseries,
title=#1, 
before upper={\scriptsize}, fontupper=\fontfamily{ptm}\selectfont,
colframe=#2, % 使用传递的参数来设定 colframe
}

\newcommand{\etc}{\emph{etc}\xspace}

\newcommand{\eg}{\emph{e.g.,}\xspace}

\title{Do we Really Need Visual Instructions? Towards Visual Instruction-Free Fine-tuning for Large Vision-Language Models}

\author{Zikang Liu\textsuperscript{1}\thanks{\:\:Equal contribution.}, Kun Zhou\textsuperscript{2}$^*$, Wayne Xin Zhao\textsuperscript{1} \thanks{\:\:Corresponding author.}, Dawei Gao\textsuperscript{3} , Yaliang Li\textsuperscript{3} , Ji-Rong Wen\textsuperscript{1}  \\
\textsuperscript{1}Gaoling School of Artificial Intelligence, Renmin University of China.\\
\textsuperscript{2}School of Information, Renmin University of China. \quad
\textsuperscript{3}Alibaba Group.\\
\texttt{\{jasonlaw8121, batmanfly\}@gmail.com, 
 francis\_kun\_zhou@163.com} \\
 \texttt{\{gaodawei.gdw, yaliang.li\}@alibaba-inc.com, jrwen@ruc.edu.cn} 
}

\begin{document}
\maketitle

\begin{abstract}
Visual instruction tuning has become the predominant technology in eliciting the multimodal task-solving capabilities of large vision-language models (LVLMs). 
Despite the success, as visual instructions require images as the input, it would leave the gap in inheriting the task-solving capabilities from the backbone LLMs, and make it costly to collect a large-scale dataset.
%synthesis of visual instructions remains costly, and training on instruction data often leads models to fit superficial vision-language response patterns, making it hard to achieve genuine capability improvements through data scaling. 
To address it, we propose ViFT, a visual instruction-free fine-tuning framework for LVLMs. 
In ViFT, we only require the text-only instructions and image caption data during training, to separately learn the task-solving and visual perception abilities.
During inference, we extract and combine the representations of the text and image inputs, for fusing the two abilities to fulfill multimodal tasks.
%also utilize the text and image as the input respectively, to extract the hidden states as the steering vectors for specific abilities, and then leverage the fused steering vector to guide LVLMs to fulfill multimodal tasks.
%Compared to conventional visual instruction tuning, LiTA only requires captions and text instructions for training, which can be scaled efficiently and effectively. Concretely, we first enable the model to learn disentangled vision and language capabilities by training on a mixture of caption and text data. Subsequently, we activate the disentangled capabilities through different modality inputs and extract task vectors from the model's hidden space. During inference, we entangle the capabilities by combining task vectors. 
Experimental results demonstrate that ViFT can achieve state-of-the-art performance on several visual reasoning and visual instruction following benchmarks, with rather less training data. 
Our code and data will be publicly released.
%Furthermore, we discover that improvements in disentangled capability obtained through data scaling can be effectively transferred to enhanced entangled vision-language capabilities for actual vision tasks, establishing a promising direction for future research.
\end{abstract}   
\section{Introduction}
\label{sec:intro}

% Through integrating visual encoder~\cite{dosovitskiy2020image,radford2021learning} with large language models~(LLMs)~\cite{brown2020language,zhao2023survey}, large vision-language models~(LVLMs)~\cite{liu2024visual,zhu2023minigpt} 
% have gained widespread attention from the research community.
% Benefiting from the remarkable instruction following ability of LLMs~\cite{ouyang2022training} and the visual perception ability of the visual encoder, LVLMs 
% can handle a variety of  multimodal tasks~\cite{fu2024mme,xiong2024llava}, and demonstrate superior performance across diverse real-world applications~\cite{du2022survey,yin2023survey}.

Recently, large vision-language models~(LVLMs), built upon existing visual encoders~\cite{dosovitskiy2020image,radford2021learning} and large language models~(LLMs)~\cite{brown2020language,zhao2023survey}, have gained widespread attention by demonstrating superior performance across diverse multimodal tasks~\cite{du2022survey,yin2023survey}. 

To empower LVLMs with multimodal task-solving capabilities, a fundamental problem is to inherit and transfer the task-solving ability of LLMs into multimodal tasks (with image inputs).
Recently, visual instruction tuning~\cite{liu2024visual,liu2024improved} has emerged as the predominant framework for achieving this goal.
Through fine-tuning on a variety of vision-language
instruction-following data from different sources, LVLMs can directly learn the corresponding knowledge and generalize into other related tasks.

Despite its success, it is still necessary to continue scaling up the number of visual instructions for fully learning multimodal advanced capabilities (\eg visual reasoning).
However, there are two bottlenecks that greatly limit the scaling of visual instructions.
%visual instruction tuning is facing two shortcomings in . 
First, due to the multimodal nature, visual instructions\footnote{Following prior works~\cite{liu2024visual}, we exclude image captions from the scope of visual instructions, as they are designed for basic vision-language alignment, instead of learning advanced multimodal task-solving capabilities.} need to incorporate visual contents (\eg images or videos) and include closely related instructions, which complicates the creation of large-scale visual instructions. Second, although existing work~\cite{liu2024visual,zhu2023minigpt} has adopted the data synthesis strategy for visual instructions, the synthesized instructions might include unreliable information regarding the visual inputs. It also poses challenges and increases the costs for quality control and scaling up. 
%Therefore, it is necessary to rethink how to efficiently learn the advanced multimodal capability
%scaling the visual instruction data is not an affordable and promising solution, to sufficiently learn the advanced . 

Considering the above challenges, we rethink whether it is feasible to reduce the reliance on visual instruction data during training LVLMs. Existing LVLMs typically map visual inputs into the LLM's token space and then generate the text output based on it. If the visual inputs are effectively perceived and aligned with text tokens, the LLM can comprehend the visual contents and leverage its inherent task-solving ability for tackling multimodal tasks. Therefore, the LVLM's multimodal task-solving capability should be the combination of (1) the visual perception ability~(for alignment) and (2) the task-solving ability from LLMs. 
% For example, when presented with a visual math problem, LVLMs should first accurately recognize image content, and subsequently apply their inherent reasoning ability to solve the problem. 
Although it is hard and costly to synthesize extensive amount of high-quality visual instructions for learning the multimodal capabilities, it is promising to sufficiently learn the two individual abilities separately, thanks to the rich resources of natural language instructions~\cite{weifinetuned,OpenHermes2-5} and image caption data~\cite{schuhmann2021laion,chen2024allava}. Therefore, our goal is to \emph{disentangle and separately strengthen} the two individual abilities during training, then \emph{combine them during inference} to enhance LVLMs.

\begin{figure}[t]
    \centering
    \includegraphics[width=0.95\columnwidth]{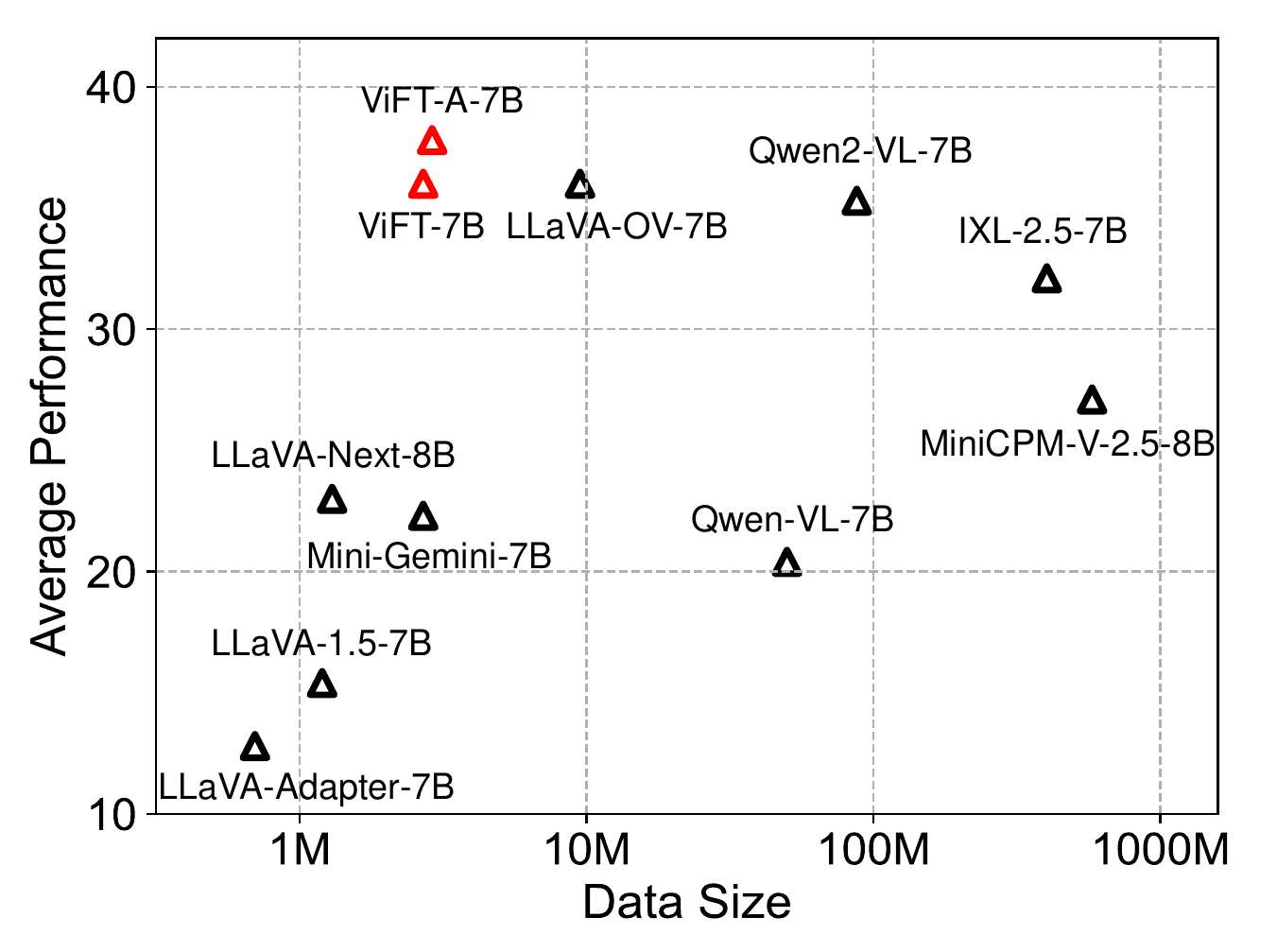}
        \caption{A comparison of ViFT with other instruction-tuned LVLM in terms of the training data size and average benchmark performance on MathVista, Mathvision, and MathVerse. ViFT is fine-tuned without any visual instruction data. For ViFT-A, we add 7\% additional simple VQA data.}
    \label{fig:intro_graph}
\end{figure}

In this work, we propose a \textbf{V}isual \textbf{I}nstruction-\textbf{F}ree fine-\textbf{T}uning framework~(ViFT) for training LVLMs.
Concretely, we need to guarantee that the two individual abilities can be independently optimized and extracted from the LVLM, and jointly elicited during inference.
To this end, we only use the image caption and natural language instruction data to jointly fine-tune the LVLM, which enhances the individual abilities by learning the image-to-text and text-to-text relations respectively, leading to less entanglement and interference. 
During inference, we extract the hidden states of the LVLM by using only the image and text parts from the input visual instruction, which are the \emph{steering vectors}~\cite{subramani2022extracting,turner2023activation} corresponding to the two individual abilities.
Through the addition of two steering vectors, the LVLM can benefit from the improvement on the individual abilities and well fulfill multimodal tasks. ViFT does not require any visual instruction data for fine-tuning, which can better inherit the original abilities from LLMs, and avoid the knowledge conflict issue caused by the divergence of visual instructions and language data.

To study the effectiveness of our approach, we conduct extensive experiments on a series of benchmarks. %Without training on any visual instructions, 
Our approach outperforms current state-of-the-art open-source LVLMs on two challenging visual reasoning benchmarks: MathVerse~\cite{zhang2025mathverse} and MathVision~\cite{wang2024measuring}. 
Compared to the best-performing LVLM, LLaVA-OneVision~\cite{li2024llava}, our ViFT achieves significant improvements on MathVerse~(34.8 vs 31.0) and MathVision~(24.0 vs 18.1) benchmarks respectively, while using less than 30\% amount of the training data, as shown in Figure~\ref{fig:intro_graph}.
The primary contributions of this work can be summarized as followed:

\begin{itemize}
    \item To the best of our knowledge, ViFT is the first instruction-free fine-tuning method with comparable performance to SOTA LVLMs.
    \item We specially designed the training and inference methods for disentangling and combining natural language task-solving and visual perception abilities, to efficiently improve the multimodal capabilities of LVLMs.
    \item Our ViFT is a low-cost approach for scaling data to improve LVLMs.
    Experimental results demonstrate the effectiveness of our approach on several benchmarks.
    % \item Based on the proposed framework, we provide a low-cost approach for scaling training data to improve LVLMs, which paves the way for future research.
\end{itemize}

\section{Related Work}
\label{sec:related-work}
\paragraph{Large Vision-Language Models.}
Large vision-language models~(LVLMs)~\cite{liu2024visual,liu2024improved} are capable of processing visual and textual inputs and tackling a variety of multimodal tasks. 
% The key to building advanced LVLM is enabling LLMs to comprehend images and inherit the LLM's task-solving abilities for multimodal tasks. 
Currently, visual instruction tuning is the predominant framework for training LVLMs. 
By training on a large number of visual instructions, LVLMs can directly learn the task-solving capabilities for the corresponding multimodal tasks. Early studies~\cite{liu2024visual,zhu2023minigpt} leverage LLMs to synthesize image-related GPT-style visual instructions. Subsequent studies leverage more advanced LVLMs~(\eg GPT-4V) for higher-quality instruction synthesis~\cite{du2023makes,chen2024allava} and quantity scaling~\cite{zhao2023svit,chen2025sharegpt4v}. 
In addition to general instruction following, another line of works focus on the LVLM's visual reasoning capability~\cite{zhang2024improve,shi2024math,gao2023g} and the performance in other visual domains~(\eg geometry~\cite{shi2024math,gao2023g}, scientific~\cite{saikh2022scienceqa}, and medical~\cite{zhang2023pmc}). Despite its success, it's costly to synthesize high-quality visual instructions, particularly when adapting to diverse new visual domains and visual tasks. 
% Furthermore, it may not be always helpful to enhance genuine capabilities through the scaling of visual instruction data~\cite{liu2024less}, as it may lead the models to learn superficial response style.

% Despite its success, visual instruction tuning tends to lead models to learn superficial response style, making it difficult to enhance genuine capabilities through the scaling of visual instruction data~\cite{liu2024less}.

\paragraph{Representation Engineering for LLMs.}
Our approach is closely related to studies of the representation engineering for LLMs~\cite{zou2023representation,turner2023activation}, which aims to extract a compact vector from the LLM's intermediate representation~(\eg hidden states). The extracted vector, also referred to as the steering vector~\cite{subramani2022extracting}, can be leveraged to manipulate the LLM's behaviour. An application of steering vectors is vector arithmetic~\cite{ilharco2022editing,turner2023activation}.
Through feature engineering~(\eg addition) of steering vectors, the LLM's behaviour can be effectively controlled.
These vectors are successfully implemented across various tasks, including style transfer~\cite{subramani2022extracting}, knowledge editing~\cite{hernandez2023inspecting}, and sentiment control~\cite{turner2023activation}. Recent researches~\cite{hendel2023context,liu2023context} extend their application to in-context learning, where they are referred to as the task vectors. In our study, we leverage steering vectors to combine the individual abilities for solving multimodal tasks.

\begin{figure*}[t]
    \centering
    \includegraphics[width=\linewidth]{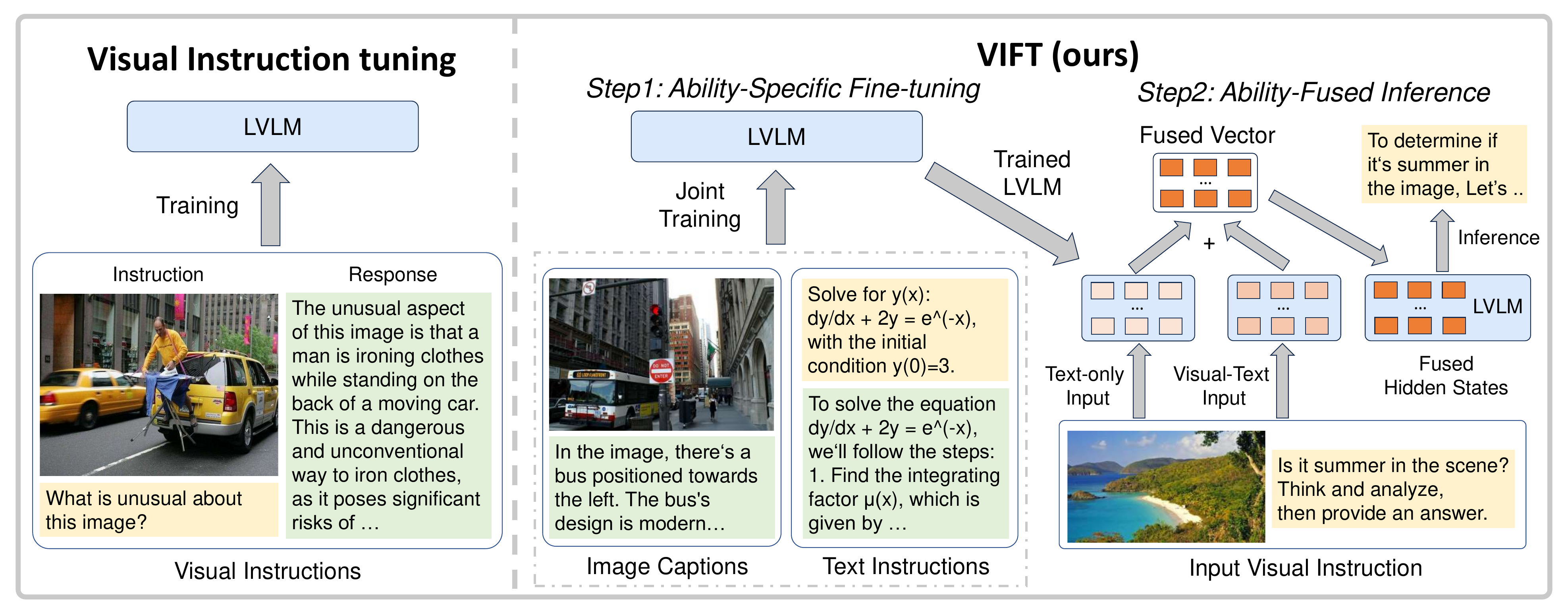}
    \caption{Compared to visual instruction tuning, ViFT first learns disentangled individual abilities through ability-specific fine-tuning. During inference, given a visual instruction, we extract the disentangled ability vectors through different modality inputs, and merge
    them into the fused vector for guiding the LVLM to generate the output.}
    \label{fig:LiTA}
\end{figure*}
\section{Preliminary}
% The motivation for training LVLM is to learn new vision capability entangle language capability to solve complex vision tasks. 
% The most commonly-used paradigm for training LVLMs is visual instruction tuning. 
%To equip LLMs with the capability to process visual information, most 
%Generally, LVLMs first encode the images, then map the visual features into LLM's input space during visual instruction tuning. We take a simple and effective LVLM design, LLaVA~\cite{liu2024improved}, as an example. LLaVA leverages 
Existing LVLMs~\cite{liu2024visual,he2024efficient} generally consist of a pretrained visual encoder $f(\cdot)$ to process visual inputs (\eg images or videos), a connection layer $g(\cdot)$ for feature projection, and an LLM $p(\cdot)$ for autoregressive generation.
During inference, given a visual instruction including an image input $v$ and a text instruction $t$, 
%and a target text response $a$. During training, 
the image is first processed through visual encoder $f(\cdot)$ and connection layer $g(\cdot)$, producing visual tokens $X_v=[x_{v_1},...,x_{v_n}]$. These tokens are then prepended to the tokens of the text input $X_t$ to compose the input of the LLM for autoregressively generating the target text. 
%For a sequence of length $L$, the probability of the target response $X_a$ is denoted as:
%\begin{align}
%p(X_a|X_v,X_t) = \prod_{i=1}^Lp(x_i|X_v, X_t, X_{a, <i})
%\end{align}
%where $X_{t,<i}$ and $X_{a, <i}$ are the instruction tokens and predicted response tokens before the current prediction token, respectively.
To train the LVLM for integrating the visual encoder and LLM, existing methods mainly incorporate two training stages: alignment pre-training and visual instruction tuning. 
The first stage only requires caption data and the second stage requires visual instructions.
% The captions are converted to the same format as visual instruction by setting the text instruction to a simple request for describing image, and the target response to the actual caption. 

In this work, we aim to skip the visual instruction tuning stage, and only train the model with captions and text instructions for disentangling and improving the task-solving and visual perception abilities.
For model architecture, we follow LLaVA's design. Concretely, we adopt SigLIP as the visual encoder according to its suggestion~\cite{liu2024llavanext}, and select Qwen2.5-7B-instruct~\cite{yang2024qwen2} as our base LLM due to its remarkable performance.
%as it's one of the most capable model among models of comparable parameter size.
For connection layer, we follow the widely-used setting in current LVLMs~\cite{liu2024visual,li2024mini} that implement a simple 2-layer MLP.
%which represents a widely accepted design paradigm .

\section{Approach}

% 能力被解耦于不同模态的输入下，模型实际上具备解决视觉任务的子能力，但是这些子能力无法在视觉-语言输入下被组合，因此仍然无法解决视觉任务。

\label{sec:approach}

%\subsection{Overview}

In this section, we introduce our proposed ViFT, a visual instruction-free fine-tuning framework for LVLMs.
Our main motivation is that the multimodal task-solving capability of LVLMs can be split into the task-solving ability of LLMs and the visual perception ability, which can be separately learned through text-only instructions and image caption data.
In ViFT, we first collect the above data to fine-tune the LVLM for learning the two individual abilities, and then extract their corresponding steering vectors to integrate the individual abilities during inference to tackle multimodal tasks.
We show the overall framework in \autoref{fig:LiTA}. 

\subsection{Ability-Specific Fine-tuning}

Previous LVLMs learn the multimodal task-solving capabilities by fine-tuning on visual instructions. In contrast, we propose to learn the task-solving and visual perception abilities separately, using text instructions and image caption data.

%disentangled vision and language capability from multimodal fine-tuning. 

\paragraph{Text Instructions.} We employ text instructions to facilitate the learning of task-solving ability. 
%For models with pre-existing language capability, these instructions help prevent the degradation of their inherent capability during multimodal training. 
Specifically, we first sample instructions from FLAN~\cite{longpre2023flan} and OpenHermes~\cite{OpenHermes2-5}. These datasets encompass a broad range of natural language tasks, including daily dialogue, knowledge utilization, multi-hop reasoning, code synthesis, \etc.
We distill responses to these queries from Qwen-2.5-72B-instruct due to its remarkable performance in multiple real-world tasks.
Additionally, we include 100K text instructions from Magpie-Qwen2.5-Pro~\cite{xu2024magpie}, which are also distilled from Qwen-2.5-72B-instruct.
%Since Qwen-2.5-72B-instruct shares the same architecture and training data with our base LLM Qwen-2.5-7B-instruct, these distilled data can be regarded as replay data~\cite{li2017learning} and better help inherit the task-solving ability for multimodal tasks.
We denote the text instruction dataset as $\mathcal{D}_{\text{text}} = \{q_i, r_i\}_{i=1}^{n_t}$, where $q_i$ and $r_i$ represent the input query and response.

\paragraph{Image Caption Data.} 
Image caption data has been widely used to improve the cross-modal alignment ability of LVLMs, enabling the models to understand and process visual inputs. We first consider the large-scale caption dataset LAION~\cite{schuhmann2021laion}, which contains a variety of web images, and sample 1M image-caption pairs from it.
%Although these web-collect captions tend to be of low quality, they are easily obtainable and can assist the model for visual perception. 
As these web-collected captions may contain low-quality noisy data, we also collect high-quality captions synthesized by GPT-4V from LLaVAR~\cite{zhang2023llavar}, ShareGPT-4V~\cite{chen2025sharegpt4v}, and ALLaVA~\cite{chen2024allava} to improve the quality of the training data.
%for better vision perception ability. 
Besides, we also collect images from specific domains~(\eg tables, graphs, documents) and caption them based on a strong LVLM, Qwen2-VL-7B~\cite{wang2024qwen2}, to enhance the visual perception ability on these types of data.
%we additionally collect images from these visual domains . 

The details of the collected visual data are presented in \autoref{apdx-train_data}. 
We denote the above caption data set as $\mathcal{D}_{\text{cap}} = \{v_i, r_i\}_{i=1}^{n_c}$, where $v_i$ and $r_i$ represent the image and caption respectively. We follow existing work~\cite{liu2024visual} to convert the caption data into instruction format to align with text instructions.
Specifically, we randomly select a caption query $q$ from a fixed query pool as its instruction. This results in a new caption dataset $\mathcal{D'}_{\text{cap}} = \{v_i, q_i, r_i\}_{i=1}^{n_c}$.

\paragraph{Simple VQA Data.} 
We find that adding a few simple VQA data is able to greatly improve the task performance. 
Thus, we collect a minimal set of simple VQA data for training,  which facilitates the LVLM in acquiring more fine-grained visual knowledge. 
Note that this kind of data is optional for training. We denote models trained with these additional VQA data as ViFT-A.

\paragraph{Training objective.} 
Following previous LVLMs, we leverage an auto-regressive training objective for optimizing the parameters within the connection layer and LLM, denoted as:
%For a LLM parameterized by $\theta$, the objective for caption data is expressed as:
\begin{align}
    \mathcal{L}(\theta) = - \sum_{j=1}^N \log \text{Pr}(r_j|v,q,r_{<j};\theta),
\end{align}
where $N$ is the target sequence length.
For text instructions, the condition of input image $v$ is given as an empty set. 
In this way, we unify the learning objectives of the two kinds of data to support joint training.
In application, due to the significant disparity in token length between captions and text instructions (as the image is converted to a long visual token sequence), we leverage a modality-specific batching strategy to prevent long padding sequences.
By separately batching the text instructions and captions, this approach can accelerate the training process while improving the disentanglement of the two individual abilities.

\subsection{Ability-Fused Inference via Steering Vectors} 

After training, the task-solving and visual perception abilities are well learned. However, they cannot be combined via standard inference. Specifically, the model will elicit each individual ability for different modality inputs, as illustrated in \autoref{apdx-cases}. We opt for the steering vectors~\cite{subramani2022extracting,turner2023activation}, which are latent vectors extracted from the model's hidden space, to address the problem. These vectors are proven to be effective for manipulating the model's behavior~\cite{subramani2022extracting}. More importantly, it enables the combination of different abilities through arithmetic operations, guiding the model to exhibit composite behavioral patterns~\cite{ilharco2022editing}. Consequently, we can activate diverse abilities through different modality inputs, extract their corresponding steering vectors, and then combine them via vector addition.

\paragraph{Extracting Steering Vector.} We focus on the LLM part of the target LVLM as it plays a crucial role in the LVLM's behavior. The LLM consists of a stack of transformer layers. During inference, the input text will be first tokenized to a sequence of tokens $\bm x = [x_1,\ldots,x_n ]$, where $n$ denotes the sequence length. Then, the sequence will be processed through multiple layers, creating intermediate hidden state vectors $\bm h^l(\bm x) = [\bm h^l(x_1),\ldots,\bm h^l(x_n)]$ at layer $l$. Notice that each input token will correspond to a hidden vector. For simplicity, we use $\bm h(\bm x)$ to denote the hidden vectors at all target layers. These hidden vectors will later be used as the steering vectors to manipulate the model's behavior.
% Previous studies~\cite{turner2023activation} discover that the model's behavior mode is closely related to its intermediate hidden states. Thus, we can extract $\bm h(\bm x)$ as the steering vector for manipulating the model's behavior.

\paragraph{Task-Solving Ability Vector.} 
Owing to our design in training, the task-solving ability is mainly learned by text-only instructions. 
Thus, we can utilize the text part of the input visual instruction to elicit the task-solving ability from the LVLM.
%We discover that without image inputs, the model's language capability is activated. As a result, the model can still exhibit normal behavior based on input text instruction. Although it does not produce the correct answer~(since no image is provided), it demonstrates a correct reasoning process. 
% As the text part is not sufficient for fulfilling the multimodal task, we only use its hidden states in LVLMs, to compose the steering vector for this ability.
Although the text part is not sufficient for fulfilling the multimodal task, it can still prompt the model to exhibit the task-solving behavior. Therefore, we aim to extract a steering vector for such ability.
Concretely, we simply use the text instruction $q$ as input, and extract the hidden vectors across all target layers.
Notably, for text-only inputs, the extracting process is the same for LLMs and LVLMs. We denote the extracted vector $\bm{h}(q)$ as the task-solving ability vector.

% and extract the activation vectors across all layers, denoted as $\bm{h}(q)$ for the task-solving ability.

\paragraph{Visual Perception Ability Vector.} 
We additionally utilize the image part of the input visual instruction, to extract the steering vector for the visual perception ability.
Here, we use the text instruction $q$ as the input, and utilize the LVLM to process the input image $v$ and text $q$. The input image and text will be converted to a sequence of tokens.
Next, we extract the hidden states of the text part from all layers, as the steering vector $\bm{h}(v, q)$.
In this way, as the text representations can attend to all image tokens, they have contained the information from the image part. Besides, they will also have the same size as the task-solving ability vector, which does not need further alignment and also supports simple fusing strategies like addition operators.

\paragraph{Ability-Fused Inference.}
After extracting the two ability vectors, we aim to combine them to activate corresponding capabilities for tackling multimodal tasks. 
Here, we devise a simple but effective ability fusion strategy via weighted addition. 
Concretely, given a visual instruction with image $v$ and text instruction $q$, the ability-fused vector is computed as:
\begin{equation}
\begin{split}
    \bm{h'}(v, q)   = \alpha \bm{h}(v, q) +  \beta \bm{h}(q) 
\end{split}
\end{equation}
Here, $\alpha$ and $\beta$ are two tunable weights. 
Given an image $v$ and a text instruction $q$, we first extract the ability vectors $\bm{h}(v, q)$ and $\bm{h}(q)$, and then compute the ability-fused vector $\bm{h'}(v, q)$.
Next, during inference, we replace the hidden representation of the input text tokens with the fused ability vector, and autoregressively generate the output tokens.
The entire generation process requires only one additional forward pass, and we will discuss the associated computational overhead in Section~\ref{sec_anaysis}.

% \begin{algorithm}[t]
% \small
% \caption{Ability-fused Inference.}
% \label{code_af_inference}
% \SetKwInOut{Input}{Input}
% \SetKwInOut{Output}{Output}

% \Input{A visual instruction $D=(v, q)$, \\ A trained model M.}
% \Output{Generated response $y$.}

% \BlankLine
% \tcp{1. Extract Task-Solving Ability Vector}
% $h(q) \leftarrow \text{ForwardPass}_M( q)$

% \BlankLine
% \tcp{2. Extract Visual Perception Ability Vector}
% $h(v, q) \leftarrow \text{ForwardPass}_M(v, q)$

% \BlankLine
% \tcp{3. Obtain Fused Ability Vector}
% $h'(v, q) = \alpha h(v,q) + \beta h(q)$

% \tcp{4. Autoregressive Generation}
% Replace representation of input text with $h'(v, q)$

% Generating response $p(y|v, t, h'(v, q)) = \prod_{i=1}^Lp(y|v, t, y_{<i}, h'(v, q))$
% \end{algorithm}
\begin{table*}[t]
  \centering
  \scalebox{0.95}{
  \begin{tabular}{>{\raggedright\arraybackslash}p{.25\linewidth}|
  *{3}{>{\centering\arraybackslash}p{.07\linewidth}}|
   >{\centering\arraybackslash}p{.07\linewidth}
   >{\centering\arraybackslash}p{.07\linewidth}|>{\centering\arraybackslash}p{.07\linewidth}
   |>{\centering\arraybackslash}p{.07\linewidth}
   }
    \toprule
    \multirow{2}{*}{Model}    & \multicolumn{3}{c}{\textbf{MathVista}} & \multicolumn{2}{c}{\textbf{MathVerse}} & \textbf{MathV}  &  \multirow{2}{*}{Avg}    \\
    \cline{2-7}
    &   MVQA   &  GVQA  &  \textbf{ALL}  &  V-mini 
 &  \textbf{ALL} &  \textbf{ALL} \\
    \midrule
    % \multicolumn{8}{c}{\textbf{Baseline LVLMs}} \\
    % \midrule
    MiniGPT4-7B   &  19.4  &   25.2   &  23.1  &  9.2   &  10.1  &  7.4  &  13.5 \\
    LLaVA-1.0-7B    &  22.4  &  28.4  &  25.2  &  10.9  & 11.5  &  8.1 &  14.9   \\
    LLaVA-1.5-7B    &  21.4  &  30.5  &  25.6  &  11.8  & 12.1  &  8.5 &  15.4   \\
    LLaVA-Next-8B     &  32.0  &  51.5  &  41.0  &  12.5  & 13.9   &  14.1  &  23.0  \\

    MiniCPM-V-2.5-8B &  39.8  &  54.5  &  46.6  &  16.8  & 20.5  &  14.1  &  27.1 \\
    LLaMA-3.2-Vision-11B  &  44.3  &  53.9  &  48.7  & 
    22.7  & 26.1  &  15.8  &  30.2 \\
    IXL-2.5-7B   &  48.0  &  \underline{62.0}  &  54.4  &  22.4  &  27.2  &  14.8  &   32.1  \\
    Qwen2-VL-7B   &  54.3  &  \textbf{64.3}  &  \textbf{58.9}  & 
 27.1  & 30.5  &  17.7  &  35.6 \\
    LLaVA-OneVision-7B    &  \textbf{58.0}  &  58.7  &  \underline{58.3}  &  26.2  & 31.0  &  18.1   &  35.8  \\
    \midrule
    % \multicolumn{8}{c}{\textbf{Ours}} \\
    % \midrule
    ViFT-7B~(ours)       &  49.8  &  48.4  &  49.2  &  \underline{28.6}  & \underline{34.8} & \textbf{24.0}   &  \underline{36.0}  \\
    ViFT-A-7B~(ours)        &  \underline{56.7}  &  53.9  &  55.4  &  \textbf{28.8}  & \textbf{35.4} & \underline{20.5}   &  \textbf{37.1} \\
    
  \bottomrule
  \end{tabular}}
  \caption{A comparison between ViFT and other baseline models on three visual reasoning benchmarks. MVQA and GVQA indicates the Math-VQA and General-VQA subset from MathVista. V-mini indicates the vision-centered subset from MathVerse. MathV is short for Math-Vision dataset. Avg indicates the average performance. \textbf{Bold} and \underline{underline} fonts indicate the best and second best performance, respectively.
  }
  \label{tab:main-table}
\end{table*}
\section{Experiment}
\label{sec:exp}

\begin{table}[t]
  \centering
  \scalebox{0.85}{
  \begin{tabular}{>{\raggedright\arraybackslash}p{.35\linewidth}
  |*{3}{>{\centering\arraybackslash}p{.12\linewidth}}|
  >{\centering\arraybackslash}p{.12\linewidth}
   }
    \toprule
    Model  &  Comp  &  Conv  &  Detail  &  ALL \\
    \midrule
    LLaVA-1.0-7B  &  66.1  &  54.7  &  44.2  &  57.2  \\
    LLaVA-1.5-7B  &  70.5  &  54.4  &  55.0  &  61.8  \\
    LLaVA-Next-8B  &  72.2  &  61.0  &  60.0  &  66.0  \\
    IXL-2.5-7B  &   78.4  &   55.6 &  \underline{74.7}  &  70.2 \\
    Qwen2-VL-7B  &   56.2  &  \textbf{87.5}  &  73.5 & 
 70.1  \\
    LLaVA-OV-7B  &  \underline{85.1}  &  \underline{79.9}  &  73.4  &  \underline{81.0}  \\
 %    Gemini-pro  &   78.8   &  \textbf{90.8}  &  67.3  & 
 % 79.9  \\
    ViFT-7B & \textbf{89.4}   &  72.3  &  \textbf{80.5}   &  \textbf{82.2} \\
    
  \bottomrule
  \end{tabular}}
\caption{A comparison between ViFT and other baseline models on visual instruction following. We report the results on LLaVABench. Comp and Conv are short for Complex and Conversation benchmark subset.}
\label{tab:table-general}
\end{table}

\subsection{Evaluation Benchmarks}

We primarily evaluate ViFT's ability on visual reasoning. Specifically, we evaluate ViFT's reasoning ability on three challenging benchmarks: MathVista~\cite{lu2023mathvista}, Mathverse~\cite{zhang2025mathverse}, and Math-Vision~\cite{wang2024measuring}.  Furthermore, we assess ViFT's general visual instruction following ability on LLaVA-Bench~\cite{liu2024visual}. Notably, for models that are only capable of generating direct answers, we employ chain-of-thought prompting to elicit its reasoning ability during visual reasoning evaluation. 
% For our evaluation framework, we employ GPT-3.5-turbo to evaluate the performance on visual reasoning, and GPT-4 to evaluate the performance on visual instruction following.

\subsection{Baselines}

% We compare ViFT with several baseline models. These models encompass early LVLMs trained with basic visual instruction data~(\eg MiniGPT4~\cite{zhu2023minigpt} and LLaVA-1.5~\cite{liu2024improved}), LVLMs trained with enhanced visual instructions~(\eg LLaVA-Next~\cite{liu2024llavanext} and MiniCPM-V-2.5~\cite{yao2024minicpm}), and current best-performing LVLMs~(\eg Qwen2-VL~\cite{wang2024qwen2} and LLaVA-OneVision~\cite{li2024llava}). 
We compare ViFT with several baseline models. These models encompass early LVLMs trained with basic visual instruction data~(MiniGPT4~\cite{zhu2023minigpt}, LLaVA-1.0~\cite{liu2024visual}, and LLaVA-1.5~\cite{liu2024improved}), LVLMs trained with enhanced visual instructions~(LLaVA-Next~\cite{liu2024llavanext}, LLaMA-3.2-Vision~\cite{meta2024llama}, and MiniCPM-V-2.5~\cite{yao2024minicpm}), and current best-performing LVLMs~(InternLM-XComposer~(IXL-2.5)~\cite{zhang2024internlm}, Qwen2-VL~\cite{wang2024qwen2}, and LLaVA-OneVision~\cite{li2024llava}). 
Further details about the baseline models are provided in the \autoref{apdx-baselines}.

% \subsection{Baselines}

% We compare LiTA with a number of existing open-source LVLMs. These models encompass early LVLMs trained with basic visual instruction data~(\eg MiniGPT4~\cite{zhu2023minigpt}, LLaVA-1.0~\cite{liu2024visual}, and LLaVA-1.5~\cite{liu2024improved}), advanced LVLM trained with enhanced visual instructions~(\eg LLaVA-Next~\cite{liu2024llavanext} and MiniCPM-V-2.5~\cite{yao2024minicpm}), and current best-performing LVLMs~(InternLM-XComposer~(IXL-2.5)~\cite{zhang2024internlm}, Qwen2-VL~\cite{wang2024qwen2} and LLaVA-OneVision~\cite{li2024llava}). Notably, Although IXL-2.5 and Qwen2-VL achieve impressive performance, they are trained on a extensive multimodal datasets (exceeding 80M samples). LLaVA-OneVision, on the other hand, utilizes a relatively smaller but still substantial dataset of 9.5M samples. Compared to these models, our LiTA demonstrates superior data efficiency by requiring only 2.7M~(2.9M for LiTA*) training samples. We present more detailed information of baseline LVLMs in \autoref{apdx-baselines}.

\subsection{Implementation Detail}

We adopt a two-stage training strategy: In the first stage, we train on web captions. In the second stage, we train on a mixture of high-quality captions and text instructions. This strategy avoids the additional computational overhead caused by significant length disparity between low-quality web captions and high-quality captions. We provide a comparison of one-stage and two-stage training in \autoref{appendix-exp}. We set the learning rate to 1e-5 for the LLM and vision encoder, and 2e-6 for the connector layer. The batch size is configured as 8 for each GPU. All models are trained for one epoch.

During inference, we only conduct ability fusion in the top 50\% of layers. We set $\alpha=1.0, \beta=0.1$ for visual reasoning and $\alpha=1.0, \beta=0.15$ for visual instruction following for robust performance. We conduct detailed studies of the optimal hyperparameters and fusion layers in Section \ref{sec_anaysis}.

\subsection{Main Results}
% 需要加入具体数据说明（哪个比哪个高，可能需要加一些paratitle，并简化文字）

\begin{table}[t]
  \centering
  \scalebox{0.9}{
  \begin{tabular}{>{\raggedright\arraybackslash}p{.50\linewidth}>{\centering\arraybackslash}p{.21\linewidth}
  >{\centering\arraybackslash}p{.24\linewidth}
   }
    \toprule
    Model  &  MathVista    &  LLaVABench  \\
    \midrule
    ViFT  &  49.2  &  82.2    \\
    $\neg$ Low-quality captions  &  48.5  &  81.6  \\
    $\neg$ High-quality captions  &  42.1  &  66.4   \\
    $\neg$ Text instructions  &  43.9  &  65.7  \\
    $\neg$ AF inference  &  46.1  &  59.1  \\
  \bottomrule
  \end{tabular}}
    \caption{The ablation of different training data component and inference strategy. AF inference indicates ability-fused inference.}
    \label{tab:table-ablation-af}
\end{table}
% We present the results of LitA on advanced visual reasoning and general visual instruction following in table xxx and table xxx, respectively. We list our findings below:

\paragraph{Visual Reasoning.} We present the results of ViFT and ViFT-A on  visual reasoning benchmarks in \autoref{tab:main-table}. Firstly, we observe that ViFT-A exhibits better performance compared to ViFT on MathVista and MathVerse, while achieving worse results on MathVision. This indicates that the VQA data indeed introduces fine-grained visual knowledge, but the performance improvement is primarily observed on benchmarks that have high visual domain overlap with these VQA data~(\eg MathVista). On benchmarks that focus more on advanced visual reasoning abilities~(\eg MathVision), the short response pattern introduced by VQA data may conflict with the model's inherited reasoning patterns, resulting in performance degradation. Overall, while the incorporation of VQA data yields a modest improvement in average performance, the enhancement remains relatively limited in scope.

Secondly, we compare ViFT with other baseline LVLMs. Notably, both ViFT and ViFT-A surpass all baseline models on MathVerse and MathVision. On MathVista, ViFT and ViFT-A show relatively lower performance compared to baselines, probably due to the lack of VQA training data. In terms of the average performance across all benchmarks,
ViFT-A demonstrates the best results among all baseline LVLMs, followed closely by ViFT. Despite being trained on a substantially smaller dataset (2.7M vs 5.5M on caption data and 0.2M vs 4M on other multimodal data), ViFT-A outperforms the leading baseline LVLM, LLaVA-OneVision-7B. This demonstrates the effectiveness of our proposed framework. Compared to conventional visual instruction tuning methods, we efficiently learn the fused visual reasoning ability to achieve superior performance with significantly less training data.

\begin{figure}[t]
  \begin{subfigure}[b]{0.48\linewidth}
  \centering
    \includegraphics[width=\linewidth]{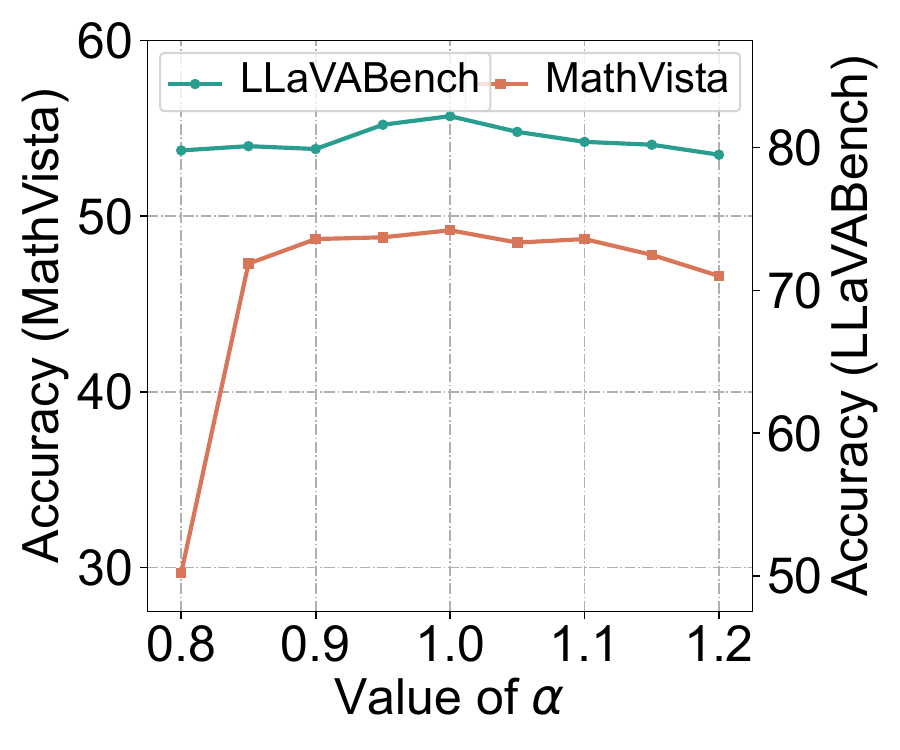}
    \caption{Impact of $\alpha$}
  \end{subfigure}\hfill
  \begin{subfigure}[b]{0.48\linewidth}
  \centering
    \includegraphics[width=\linewidth]{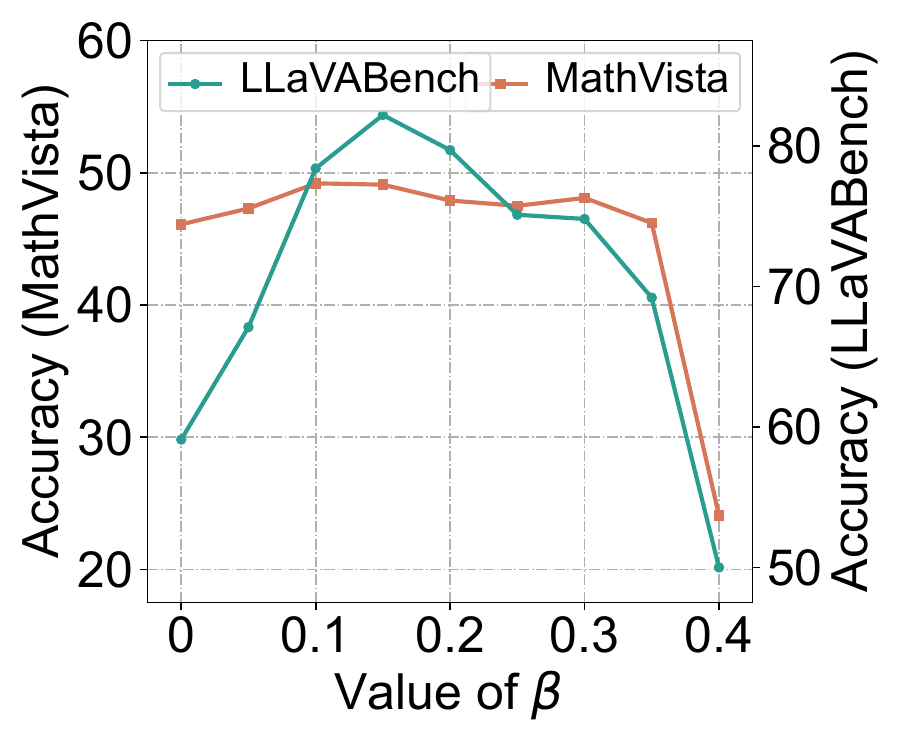}
    \caption{Impact of $\beta$}
  \end{subfigure}
  \caption{The impact of different hyperparameters.}
  \label{tab:ablation-alpha-beta}
\end{figure}

\paragraph{Visual Instruction Following.} 
We demonstrate the performance of ViFT on general visual instruction following in \autoref{tab:table-general}. Among all baseline models, LLaVA-OneVision-7B exhibits the best average performance, demonstrating promising results across all subsets. We also observe the poor performance of Qwen2-VL-7B on the complex subset. Given that it employs Qwen2-7B, a strong backbone LLM, such low performance likely stems from the compromised instruction following ability caused by visual instruction tuning.
Compared to baseline models, ViFT achieves the best average performance. As for each subset, ViFT has the best performance on the complex subset, indicating that our approach enables the model to handle complex instructions while correctly interpreting the images. ViFT also has the strongest performance on the detail subset, which can be credited to the high-quality caption data. Despite this, ViFT's performance on the conversation subset is relatively mediocre, suggesting that the model can be further improved by conducting human alignment based on visual inputs.

\section{Further Analysis}\label{sec_anaysis}
\paragraph{Ablation Study.}

We employ diverse training data components and inference strategy in our training framework. We present the ablation results in \autoref{tab:table-ablation-af}.
First, we examine the impact of each data component for fine-tuning. We observe that removing high-quality captions or text instructions can result in severe performance decline. This indicates that the high-quality captions play a crucial role in enhancing the model's visual perception ability, which subsequently improves their multimodal task-solving capability. Text instructions are equally important as they preserve the LLM's inherent task-solving ability from multimodal training. In comparison, the impact of low-quality captions is relatively limited. Second, we study the effect of our proposed ability-fused inference. As we can observe, the model exhibits significant performance degradation without ability-fused inference.  This indicates that the individual abilities acquired through fine-tuning cannot be effectively combined through standard inference, and our proposed ability-fused inference successfully addresses this limitation.

\paragraph{Hyperparameter Tuning.}
We study the effect of different hyperparameter $\alpha$ and $\beta$ on model performance. The results are presented in \autoref{tab:ablation-alpha-beta}. 
For $\alpha$, we observe that as $\alpha$ increases, the model's performance initially increases and then decreases. 
While the performance on MathVista exhibits a sudden improvement at early stages, it remains relatively stable as $\alpha$ changes. The results confirm that $\alpha=1.0$ represents an optimal choice, while small deviations do not significantly impact performance.
Similar to  $\alpha$, the model performance exhibits an increase-then-decrease pattern as $\beta$ varies, though with more pronounced fluctuations. We observe a sudden performance drop when $\beta$ reaches $0.4$, indicating that such a large $\beta$ can result in the model's abnormal behavior. Furthermore, we discover that the optimal $\beta$ varies across different tasks. For visual reasoning, the optimal $\beta$ is 0.1, whereas for visual instruction following, it is 0.15. This demonstrates that different vision tasks may require varying levels of individual abilities, resulting in task-specific optimal fusion ratios.

\begin{table}[t]
  \centering
  \scalebox{0.9}{
  \begin{tabular}{>{\centering\arraybackslash}p{.18\linewidth}
  |*{2}{>{\centering\arraybackslash}p{.21\linewidth}}
  >{\centering\arraybackslash}p{.24\linewidth}
   }
    \toprule
    Layers  &  MathVista  &  MathVision  &  LLaVABench  \\
    \midrule
    0-7  &  48.8  &  17.0  &  64.8  \\
    0-14  &  46.4  &  19.6  &  74.4  \\
    0-21  &  \underline{49.0}  &  17.3  &  80.6  \\
    0-28  &  47.8  &  16.4  &  76.8  \\
    7-28  &  48.1  &  19.6  &  \underline{81.2}  \\
    14-28  &  \textbf{49.2}  &  \textbf{24.0}  &  \textbf{82.2}  \\
    21-28  &  46.3  &  \underline{22.8}  &  73.8  \\
  \bottomrule
  \end{tabular}}
    \caption{The impact of fusion layer selection.}
    \label{tab:table-ablation-fl}
\end{table}

\paragraph{Fusion Layer Selection Analysis.}

We investigate the impact of layer selection for ability fusion. We examine two strategies: selecting layers from the top downward, or from the bottom upward. The results are presented in \autoref{tab:table-ablation-fl}. Our findings indicate that the top-down selecting strategy consistently outperforms bottom-up selection when selecting the same number of layers. This is likely due to the fact that the LLM's top layers have more influence on the model's generation behavior~\cite{geva2020transformer,geva2022transformer}, which makes ability fusion more effective at these layers. Also, recent studies~\cite{chen2025image,zhang2025llava} demonstrate that visual information tends to aggregate with text tokens within the LVLM's early layers, and ability fusion in these layers may disrupt such a process, leading to declined performance. Moreover, we discover that selecting 50\% of the layers from the top of the model downward yields the best performance, which makes it an optimal choice.

\begin{figure}[t]
  \begin{subfigure}[b]{0.48\linewidth}
  \centering
    \includegraphics[width=\linewidth]{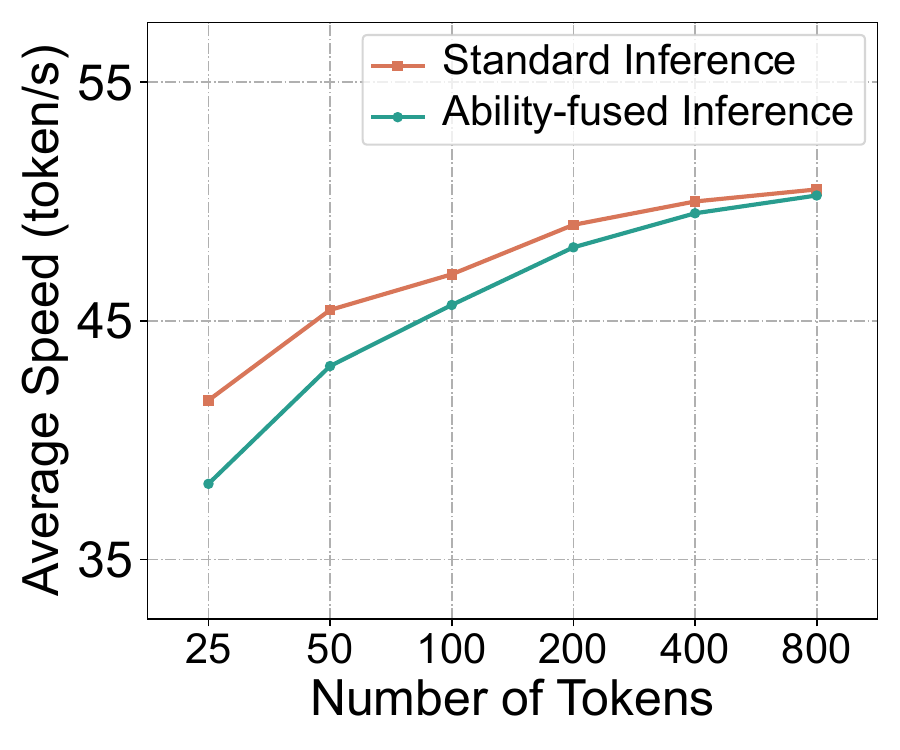}
    \caption{Efficiency Test}
    \label{fig:speed}
  \end{subfigure}\hfill
  \begin{subfigure}[b]{0.48\linewidth}
  \centering
    \includegraphics[width=\linewidth]{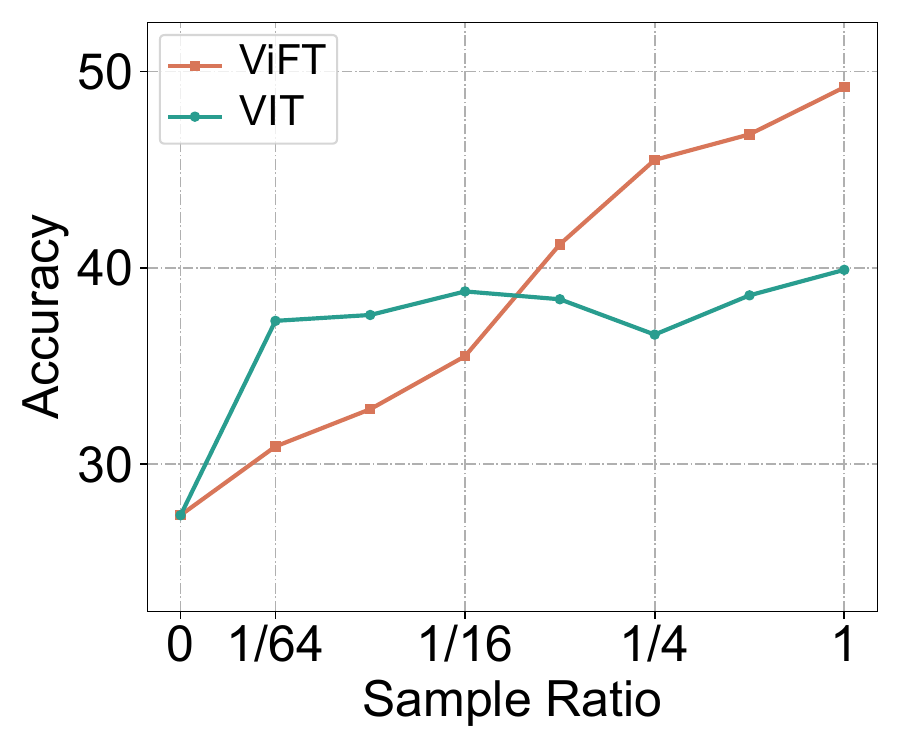}
    \caption{Scaling Test}
    \label{fig:scaling}
  \end{subfigure}
  \caption{Efficiency test and scaling test for ViFT.}
\end{figure}

\paragraph{Computation Complexity.}
We examine the additional time overhead of ability-fused inference compared to standard inference. The results are presented in \autoref{fig:speed}.
As we can observe, when generating short responses~(\eg 25 tokens), our ability-fused inference is about 8\% slower than standard inference. However, as the generation length increases, the speeds for standard inference and ability-fused inference gradually converge. When generation length reaches 400, ability-fused inference almost doesn't introduce any additional computational overhead. This aligns with our expectations. For ability-fused inference, we merely introduce one additional forward pass during the entire generation process. Thus, while there is some discrepancy when generating short responses, such differences become negligible as generation length increases.

% \begin{figure}[t]
%     \centering
%     \includegraphics[width=0.95\columnwidth]{images/speed_v2.pdf}
%     \caption{Speed comparison between standard inference and our proposed debias inference.}
%     \label{fig:speed}
% \end{figure}

\paragraph{Scaling Test.}
We investigate the effect of data scaling for conventional visual instruction tuning~(VIT) and ViFT. Concretely, we collect visual instruction datasets from ALLaVA~\cite{chen2024allava}, SViT~\cite{zhao2023svit} and LLaVA-Instruct~\cite{liu2024visual}. Then, we randomly sample several data subsets from ViFT's training data and the collected visual instructions at different sampling ratios, respectively. We then train LVLMs with these data subsets and evaluate their results on MathVista, as shown in \autoref{fig:scaling}. We observe that data scaling consistently yields performance improvements for ViFT. This indicates that the enhancements in two individual abilities effectively propagate to improved fused multimodal task solving capability.
As for conventional visual instruction tuning, the model achieves promising performance improvement with minimal data, but cannot yield significantly better results via data scaling. This likely occurs because existing visual instructions primarily help models learn superficial styles, rather than improving actual multimodal task-solving capability. This further validates that ViFT demonstrates greater potential for performance improvement by leveraging existing large-scale, cost-effective data, compared to conventional approaches.

% \begin{figure}[t]
%     \centering
%     \includegraphics[width=0.95\columnwidth]{images/scaling_test_v2.pdf}
%     \caption{Comparison of LiTA and VIT~(Visual Instruction Tuning) on data scaling.}
%     \label{fig:scaling}
% \end{figure}

% \begin{figure}[t]
%   \includegraphics[width=0.48\linewidth]{images/speed_v2.pdf}
%   \hfill
%   \includegraphics[width=0.48\linewidth]{images/scaling_test_v2.pdf}
%   \caption {The ablation of $\alpha$ and $\beta$.}
%   \label{tab:ablation-alpha-beta}
% \end{figure}

\section{Conclusion}
\label{sec:conclusion}

In this paper, we proposed an instruction-free fine-tuning framework ViFT, for enhancing the multimodal task-solving capabilities of LVLMs. Concretely, instead of using visual instructions, we only leveraged text instructions and image caption data, to separately learn the individual task-solving and visual perception abilities for the LVLM.
%vision and language capability via disentangled fine-tuning. 
After that, we extracted the steering vectors by using the model's hidden space for the disentangled abilities via different modality inputs, and combined them to guide the inference of the LVLM in multimodal tasks.
%During inference, we entangled the capabilities by combining the task vector in prefilling stage. 
With rather less training data, our trained model, ViFT, achieved state-of-the-art performance among competitive LVLMs across various visual reasoning and instruction following benchmarks. Furthermore, based on our proposed framework, we can efficiently scale the vision data and text data to enhance the model's performance, which facilitates further advancements in this field.

% In this paper, we propose a new visual instruction-free fine-tuning framework LiTA, for training LVLMs. Concretely, we 

% we study the modality bias in existing large vision-language models. We propose a new training paradigm, LiTA, to mitigate the modality bias in LVLMs. Concretely, we first remove all visual instructions from the training data to avoid overfitting to the instruction-response pattern. Then, we propose to extract a disentangled latent vector that causes the biased behaviour. After that, we calibrate the LVLM's behaviour by arithmetically subtracting the extracted latent vector from instruction hidden states. Our training strategy only require caption and text for training, and can effectively transfer LLM's advanced capabilities to visual scenarios. LiTA achieves state-of-the-art performance on several challenging visual reasoning benchmark, surpassing current best-performing LVLMs that is trained on much more complex visual instruction data.

\section{Limitations}

In this paper, we propose ViFT, a visual instruction-free fine-tuning framework for training LVLMs. While our approach achieves promising performance on downstream benchmarks, it still has some potential limitations. First, we prove that our approach can be enhanced by scaling the vision data and text data for training. Since there already exists well-established methods for efficiently synthesizing such data in large quantities~\cite{yu2023metamath,zhou2024jiuzhang3}, there is still room for further improvement. Second, we utilize captions as the primary multimodal data to facilitate the learning of visual perception ability. While this approach represents the current mainstream practice, whether coarse-grained captions constitute the optimal data choice for visual perception learning across all vision domains remains an open research question. Third, apart from the visual reasoning and instruction following capabilities, ViFT has the potential to efficiently transfer more advanced capabilities to visual tasks~(\eg long-thought reasoning~\cite{openai2024o1,guo2025deepseek}), and we will further explore this direction in subsequent work.

\newpage
\bibliography{custom}

\newpage

\appendix
% \section{Experimental Details}\label{sec:appendix}

\section{Training Dataset}\label{apdx-train_data}
We utilize OpenHermes~\cite{OpenHermes2-5} and FLAN~\cite{longpre2023flan} as sources for text instruction queries. Open-Hermes comprises a diverse collection of text instructions from various sources and FLAN contains a substantial set of task-specific instructions. We anticipate these instruction sets will enhance the model's language capability in both general scenarios and complex reasoning tasks. Following query acquisition, we employ Qwen2-72B-instruct to distill specific instruction responses. This approach is adopted because the distilled responses demonstrate higher quality compared to the original responses. Additionally, since Qwen2-72B-instruct shares the same training data as our base LLM (though with different parameter scales), we hypothesize this alignment would better preserve the original language capabilities.

As for vision data, we collect extensive caption datasets encompassing both general and domain-specific vision domains. Additionally, we incorporate a few simple VQA datasets to enhance the model's learning of vision knowledge. The details of the vision data are presented in \autoref{tab:table-caption-statistic} and \autoref{tab:table-vqa-statistic}.

\section{Evaluation Datasets}\label{evaluation_dataset}
We evaluate ViFT on four downstream benchmarks, the details of the benchmarks are as followed:

\begin{itemize}
    \item \textit{MathVista:}~\citep{lu2023mathvista} it evaluates the LVLM's mathematical reasoning capabilities in multiple vision domains. It contains 6141 evaluation data samples, collected from 28 existing datasets and 3 newly created datasets. 
    \item \textit{MathVerse:}~\citep{zhang2025mathverse} it is an in-depth benchmark for evaluating LVLM's reasoning capability. It consists of 2612 math problems, and each problem is transformed into 6 distinct problem versions. We report the full performance~(ALL) and the performance on vision-mini~(V-mini) subset in our experiments. 
    \item \textit{MathVision:}~\citep{wang2024measuring} it develops a comprehensive and challenging benchmark for 
    evaluating the LVLM's advanced reasoning skill. It comprises 3040 high-quality mathematical problems derived from authentic mathematics competitions. These problems encompass several distinct mathematical disciplines and are categorized across five difficulty levels.
    
    \item \textit{LLaVABench:}~\citep{liu2024visual} t evaluates the model's instruction-following capabilities across diverse visual scenarios, assessing its conversation, detailed description, and complex reasoning skills. Through carefully designed prompts, it measures the model's ability to generate accurate responses based on complex instructions in visual contexts.

\end{itemize}

Following previous studies~\cite{zhang2024internlm,li2024llava}, we conduct evaluations on the {\emph{testmini}} set of MathVista and MathVision. For our evaluation framework, we employ GPT-3.5-turbo to evaluate the performance on visual reasoning, and GPT-4 to evaluate the performance on visual instruction following.

\begin{table}[t]
  \centering
  \scalebox{0.9}{
  \begin{tabular}{
  >{\centering\arraybackslash}p{.28\linewidth}
  *{2}{>{\centering\arraybackslash}p{.24\linewidth}
  }}
    \toprule
    Image Source &  Domains   &  Numbers  \\
    \midrule
    ALLaVA  & General  &  564976      \\
    LLaVAR  &  General   &  250000 \\
    ShareGPT    &  General   &  200000 \\
    COCO  &  General   &  82783  \\
    VG &  General  &  158557  \\ 
    Synthdog & OCR  &  29765  \\
    GeoQA  &  Math   &   6027   \\
    CLEVR  &  Math   &  20000 \\
    FigureQA  &  Figure  &  20000 \\
    DocVQA  &  Document  &  10194  \\
    TabMWP   &   Table   &  20000 \\
    ChartQA  &  Chart   &  18317  \\
    DVQA  &  Diagram   &  30000 \\
    MMarxiv  &  Academic  &  54399 \\
    IconQA  &  Icons  &  18946  \\
    AI2D   &  Science  &  4903  \\
    ScienceQA   &  Science  &  6757  \\

  \bottomrule
  \end{tabular}}
    \caption{Details of caption training data.
    \label{tab:table-caption-statistic}
  }
\end{table}

% \begin{figure}[t]
%     \centering
%     \includegraphics[width=0.95\linewidth]{images/prompts.pdf}
%     \caption{Prompts for caption generation.}
%     \label{fig:prompts}
% \end{figure}

\begin{table}[t]
  \centering
  \scalebox{0.9}{
  \begin{tabular}{
  >{\centering\arraybackslash}p{.29\linewidth}
  *{2}{>{\centering\arraybackslash}p{.24\linewidth}
  }}
    \toprule
    Image Source &  Domains   &  Numbers  \\
    \midrule
    VQAv2  &  General   &  20000  \\
    GQA &  General  &  20000  \\ 
    AI2D   &  Science  &  12413  \\
    CLEVR  &  Math   &  20000 \\
    CLEVR-Math  &  Math   &  20000 \\
    DocVQA  &  Document  &  10194  \\
    FigureQA  &  Figure  &  20000 \\
    Geometry3K  &  Math  &  2101 \\
    GeoQA  &  Math  & 20000  \\
    IconQA  &  Icons  &  20000  \\
    PlotQA  &  Figure   &  20000 \\ 
    DocVQA  &  Document  &  10194  \\
    ChartQA  &  Chart   &  18317  \\
    DVQA  &  Diagram   &  20000 \\
  \bottomrule
  \end{tabular}}
    \caption{Details of VQA training data~(optional).
    \label{tab:table-vqa-statistic}
  }
\end{table}

\section{Baselines}\label{apdx-baselines}

We compare ViFT with a number of existing open-source LVLMs. 
% These models include early LVLMs~(MiniGPT4~\cite{zhu2023minigpt}, LLaVA-1.0~\cite{liu2024visual}, and LLaVA-1.5~\cite{liu2024improved}), advanced LVLM trained with better-quality visual instructions~(LLaVA-Next~\cite{liu2024llavanext}, LLaMA-3.2-Vision~\cite{meta2024llama}, and MiniCPM-V-2.5~\cite{yao2024minicpm}), and current state-of-the-art LVLMs~(InternLM-XComposer~(IXL-2.5)~\cite{zhang2024internlm}, Qwen2-VL~\cite{wang2024qwen2} and LLaVA-OneVision~\cite{li2024llava}). 
Notably, Although IXL-2.5 and Qwen2-VL achieve impressive performance, they are trained on a extensive multimodal datasets (exceeding 80M samples). LLaVA-OneVision, on the other hand, utilizes a relatively smaller but still substantial dataset of 9.5M samples. Compared to these models, our ViFT demonstrates superior data efficiency by requiring only 2.7M~(2.9M for ViFT-A) training samples. We present more detailed information of baseline LVLMs in \autoref{apdx-baselines}.
Apart from baseline LVLMs introduced above, we also include LLaVA-Adapter~\cite{zhang2024llama} and Mini-Gemini~\cite{li2024mini} in \autoref{fig:intro_graph}. We report the training data size of these models in \autoref{tab:table-statistic}. For models without exact number of training data size in papers, we estimate the lower bound of the data size. For IXL-2.5, we report the incomplete training data size in their paper. For Qwen2-VL, we estimate the training data size by dividing the total training tokens with the max token length for each sample.

\begin{table}[t]
  \centering
  \scalebox{0.9}{
  \begin{tabular}{>{\centering\arraybackslash}p{.33\linewidth}
  |>{\centering\arraybackslash}p{.22\linewidth}
  >{\centering\arraybackslash}p{.18\linewidth}
  >{\centering\arraybackslash}p{.17\linewidth}
   }
    \toprule
    Model  & \# Captions  &  \# Others &  \# Total  \\
    \midrule
    LLaVA-1.5  &  560K  &  665K  &  1.2M  \\
    LLaVA-Next  &  560K  &  760K  &  1.3M  \\
    MiniCPM-V-2.5  &  570M  &  8.3M  & 578M   \\
    IXL-2.5  &  > 400M  &  > 2M  &  > 402M   \\
    Qwen2-VL  &  > 87.5M  &  -  &  > 87.5M  \\
    LLaVA-OV   &  5.5M  &  4.0M  &  9.5M  \\
    ViFT  &  2.7M  &  0     &  2.7M  \\
    ViFT-A  &  2.7M  &  200K   &  2.9M  \\
    
  \bottomrule
  \end{tabular}}
    \caption{The statistics of training data for ViFT and other baseline LVLMs. Given that Qwen2-VL and IXL do not provide their specific training data volumes, we estimate the minimum data size based on the training data descriptions provided in their  papers.}
    \label{tab:table-statistic}
\end{table}

% \section{Implementation Detail}
% Ideally, our approach could be completed in a single training stage. However, due to the significant length disparity between low-quality web captions and high-quality captions, aligning their lengths during training would introduce additional computational overhead. Therefore, we adopted a two-phase training strategy: In the first stage, we train on web captions. In the second stage, we train on a mixture of high-quality captions and text instructions. This training approach effectively reduces time complexity without compromising performance. During fine-tuning, the learning rate is set to 1e-5 for the LLM and 2e-6 for the projector. The batch size is configured as 8 for each GPU, All models are trained for one epoch.

% During inference, we omitt the entanglement process in the first 25\% of layers. This decision is supported by previous studies~\cite{chen2025image} demonstrating that visual information aggregates with instruction tokens in the early layers of the model. We aim to preserve this visual information aggregation process to maintain robust vision capability. We set different $\beta$ for different tasks to achieve optimal performance. Concretely, we set $\beta=0.05$ for visual reasoning tasks and $\beta=0.1$ for visual instruction following tasks.

\section{Additional Experiments}\label{appendix-exp}
\subsection{One-stage training vs two-stage}
During training, we adopt a two-stage training strategy to reduce cost. We investigate whether a two-stage training strategy produces significant different results compared to a one-stage approach. The results are presented in \autoref{tab:table-train}. As we can observe, single-stage training exhibits almost the same performance as two-stage training, yet is more efficient.

% \subsection{The Effect of Ability-fused Inference}

% We investigate the effectiveness of our proposed ability-fused inference. Specifically, we set both $\alpha$ and $\beta$ to 0 and evaluate the model's performance. The results are shown in \autoref{tab:table-ablation-afi}. We observe that without ability fusion, the model exhibits significant performance degradation across all benchmarks. Specifically, during standard inference, the model merely generates image captions rather than following instructions, as shown in \autoref{tab:case1} and \autoref{tab:case2}. This indicates that the individual abilities acquired through fine-tuning cannot be effectively combined through standard inference to address practical visual tasks. However, our proposed ability-fused inference successfully addresses this limitation.

\begin{table}[t]
  \centering
  \scalebox{0.9}{
  \begin{tabular}{
  >{\centering\arraybackslash}p{.22\linewidth}
   >{\centering\arraybackslash}p{.15\linewidth}
  >{\centering\arraybackslash}p{.18\linewidth}
  >{\centering\arraybackslash}p{.12\linewidth}
  >{\centering\arraybackslash}p{.12\linewidth}
   }
    \toprule
    Strategy   & Time & MathVista  &  LBench  &  MathV  \\
    \midrule
    Two-Stage  &  $\approx$ 36h  &  49.2  &  24.0 &  82.2    \\
    One-Stage  &  $\approx$ 47h &   49.8  &  23.2  &  81.9   \\

  \bottomrule
  \end{tabular}}
    \caption{Comparison of ViFT with different training strategy.
  }
  \label{tab:table-train}
\end{table}

% \begin{table}[t]
%   \caption{The ablation of ability-fused inference. $\neg$ AF indicates inference without ability fusion.}
%   \label{tab:table-ablation-afi}
%   \centering
%   % \renewcommand\arraystretch{1.1}
%   \scalebox{0.9}{
%   \begin{tabular}{>{\centering\arraybackslash}p{.18\linewidth}
%   |*{2}{>{\centering\arraybackslash}p{.21\linewidth}}
%   >{\centering\arraybackslash}p{.24\linewidth}
%    }
%     \toprule
%     Model  &  MathVista  &  MathVision  &  LLaVABench  \\
%     \midrule
%     ViFT  &  49.2  &  24.0 &  82.2   \\
%     $\neg$ AF  &  46.1 &  13.5  &  59.1   \\
%   \bottomrule
%   \end{tabular}}
% \end{table}

\section{Case Studies}\label{apdx-cases} 
This section presents ViFT's generating behaviour under various experimental settings, and a comparison of ViFT with other baseline LVLMs. The detailed results are presented below.

\paragraph{Comparison of ViFT with Existing LVLMs.}  We present the results of ViFT and other SoTA LVLMs in \autoref{tab:case5}, \autoref{tab:case10}, and \autoref{tab:case11}. As we can observe, many state-of-the-art baseline LVLMs fail to  generate a high-quality reasoning path even when chain-of-thought prompt is leveraged, such as Qwen2-VL and LLaVA-OneVision. LLaMA-3.2-Vision-11B is the only baseline LVLM that is capable of generating relatively coherent chains of thought, but it may occasionally produce minor error during the reasoning process~(\eg misinterprets the sine function), which results in an erroneous final result. InternLM-XComposer-2.5 sometimes generates promising reasoning path, but at other times produces shorter responses. For Qwen2-VL-7B and LLaVA-OneVision-7B, their reasoning is extremely short, leading to incorrect results. As this is actually a simple geometric problem, which should not pose a challenge for LLMs, the observed erroneous behavior is likely due to a ability degradation resulting from conflicts between knowledge learned from visual instruction tuning and the intrinsic abilities of LLM itself. Compared to these models, our ViFT is the only one that can almost fully inherit the original reasoning capabilities from its backbone LLM, generate high-quality reasoning paths, and ensure the accuracy and consistency of intermediate steps, ultimately achieving more correct results.

\paragraph{ViFT's Behaviour With or Without Ability-fused inference.}

We present the results of ViFT with or without ability-fused inference in \autoref{tab:case0}, \autoref{tab:case1}, and \autoref{tab:case2}. First, we discover that the fine-tuned model will elicit diverse individual ability with different modality inputs, instead of a combined capability via standard inference. Specifically, it mostly produces a image description for multimodal inputs, regardless of the actual instruction. However, when we leverage the same instruction part as text-only input, the model follows the instruction and exhibits expected behavior, though it fails to provide a correct answer due to the missing image. Second, after implementing our ability-fused inference, ViFT maintains its visual perception capability while successfully leveraging the fused task-solving ability to generate correct responses, validating the effectiveness of our method.

% In \autoref{tab:case0}, we present the results of ViFT without ability-fused inference. We discover that after ability-specific fine-tuning, the model can not elicit normal multimodal task-solving ability during standard inference and will elicit diverse behavior with different modality inputs. Specifically, it mostly produces a description for the given image, regardless of the actual instruction. However, when we leverage the same instruction without the image as input, the model successfully follows the given instruction and exhibits expected behavior, though it fails to provide a correct answer due to the missing image.

% In \autoref{tab:case1} and \autoref{tab:case2}, we present the results of ViFT with or without ability-fused inference. We discover that ViFT tends to generate captions without ability fusion. In \autoref{tab:case1}, ViFT generates a caption first, then directly produce a incorrect answer without actually reasoning. In \autoref{tab:case2}, ViFT merely generates a caption and doesn't answer the question at all. The generated captions are accurate, demonstrating ViFT's effective visual perception ability. However, the lack of task-solving ability results in incorrect behavior. Nevertheless, after implementing our ability-fused inference approach, ViFT maintains its visual perception capability while successfully leveraging the fused task-solving ability to generate correct responses, validating the effectiveness of our method.

\paragraph{ViFT's Behaviour With Different $\alpha$ Setting.} We present the results of ViFT's behaviour with different $\alpha$ in \autoref{tab:case4}. We observe that when $\alpha$ is set to a low value~(0.8 in our case), ViFT rejects to answer the question for lack of visual information. This may be attributed to diminished visual perception ability caused by low value of $\alpha$. As $\alpha$ gradually increases, the model begins to exhibit normal reasoning behaviour and generates accurate results. Within a certain range of $\alpha$, the model's behavior remains notably consistent, showing no significant variations despite the changes of $\alpha$ values. We also notice that when $\alpha$ reaches $1.1$, the model starts to first generate a image description, then answer the question. While the reasoning behavior remains largely consistent, this indicates that a larger $\alpha$ may enhance the model's focus on visual information.

\paragraph{ViFT's Behaviour With Different $\beta$ Setting.}

The results of ViFT's behaviour with different $\beta$ are presented in \autoref{tab:case3}. We discover that when $\beta=0$, the model exhibits a caption-only behaviour, similar to ViFT without ability fusion. As $\beta$ increases, the model progressively demonstrates consistent reasoning behavior, aligning with the behavioral patterns observed during $\alpha$ variation. However, when $\beta$ reaches $0.4$, ViFT will output random strings, which indicates that a large $\beta$ can result in unstable behaviour in the model. This also corresponds to the previous findings where the model's performance experiences a sudden and drastic decline once the beta value reaches a certain threshold.

\begin{table*}[htbp]
    \centering
    \renewcommand{\arraystretch}{1.2} 
    \setlength{\tabcolsep}{8pt} 
    \begin{tabular}{p{14cm}}
        \centering
        \includegraphics[width=0.6\textwidth]{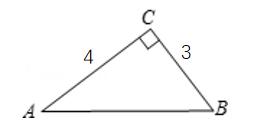}
        
        \begin{tcolorbox}[colback=gray!5, colframe=gray!40, sharp corners, width=\linewidth, boxrule=0.5mm]
        \small
        \textbf{Problem: }
        Please first conduct reasoning, and then answer the question and provide the correct option letter, e.g., A, B, C, D, E, at the end.
        
        As shown in the figure, then the value of sinB is equal to ()
        Choices:
        
        A:4/3
        
        B:3/4
        
        C:4/5
        
        D:3/5

        \textbf{Ground Truth:} C
        \end{tcolorbox}  % Reduce the vertical space between input and output
        \begin{tcolorbox}[colback=gray!5, colframe=gray!40, sharp corners, width=\linewidth, boxrule=0.5mm]
            \small
            \textbf{LLaMA-3.2-Vision-11B:}
            \\
**Step 1: Understand the problem**

The problem asks for the value of sinB in the given figure. To find this, we need to use trigonometric concepts, specifically the definition of sine in a right-angled triangle.

**Step 2: Identify the triangle and its sides**

From the figure, we can see that triangle ABC is a right-angled triangle with sides labeled as follows:

*   Side AB = 3
*   Side BC = 4
*   Hypotenuse AC = 5

**Step 3: Recall the definition of sine**

The sine of an angle in a right-angled triangle is defined as the ratio of the length of the side opposite the angle to the length of the hypotenuse.

**Step 4: Apply the definition of sine to find sinB**

In this case, angle B is opposite side AB, and the hypotenuse is AC. Therefore, sinB = AB / AC.

**Step 5: Calculate sinB**

Using the values from step 2, we can calculate sinB as follows:

sinB = AB / AC = 3 / 5

Answer: D

        \end{tcolorbox}

        \begin{tcolorbox}[colback=gray!5, colframe=gray!40, sharp corners, width=\linewidth, boxrule=0.5mm]
        \small
        \textbf{InternLM-XComposer-2.5-7B:}
        \\
To find the value of sinB, we need to use the definition of sine in a right-angled triangle. The sine of an angle in a right-angled triangle is the ratio of the length of the side opposite the angle to the length of the hypotenuse. In this case, side AC is opposite angle B, and AB is the hypotenuse.

Given that AC = 3 and AB = 4, we can calculate sinB as follows:

sinB = opposite/hypotenuse
sinB = 3/4

Therefore, the correct answer is B: \(\frac{3}{4}\).

\end{tcolorbox}

        \begin{tcolorbox}[colback=gray!5, colframe=gray!40, sharp corners, width=\linewidth, boxrule=0.5mm]
        \small
        \textbf{Qwen2-VL-7B:}
        \\
Solution: In right triangle triangle ABC, angle C=90°, AC=3, BC=4.
According to the Pythagorean theorem, we have AB=5.
Therefore, sinB=AC/AB=3/5.
So, the answer is D.

\end{tcolorbox}

\end{tabular}
\caption{A comparison of ViFT with other baseline LVLMs~(part1).}
\label{tab:case5}
\end{table*}

\begin{table*}[htbp]
    \ContinuedFloat
    \begin{tabular}{p{14cm}}
        \centering

                \begin{tcolorbox}[colback=gray!5, colframe=gray!40, sharp corners, width=\linewidth, boxrule=0.5mm]
        \small
        \textbf{LLaVA-OneVision-7B:}
        \\
Solution: Since angle C=90°, we have sinB=AC/AB=3/5.
Therefore, the answer is D.

\end{tcolorbox}
        \begin{tcolorbox}[colback=gray!5, colframe=gray!40, sharp corners, width=\linewidth, boxrule=0.5mm]
        \small
        \textbf{ViFT-7B:}
        \\
To determine the value of sinB, we need to identify the sides of the triangle relative to angle B. The sine of an angle in a right-angled triangle is defined as the ratio of the length of the side opposite the angle to the length of the hypotenuse.

From the diagram:
- Side AC (opposite to angle B) has a length of 4 units.
- Side BC (adjacent to angle B) has a length of 3 units.
- Side AB (the hypotenuse) is not explicitly given but can be calculated using the Pythagorean theorem if needed.

Using the Pythagorean theorem for triangle ABC:
\[ AB^2 = AC^2 + BC^2 \]
\[ AB^2 = 4^2 + 3^2 \]
\[ AB^2 = 16 + 9 \]
\[ AB^2 = 25 \]
\[ AB = 5 \]

Now, we can calculate sinB:
\[ \sin B = \frac{\text{opposite}}{\text{hypotenuse}} = \frac{AC}{AB} = \frac{4}{5} \]

Therefore, the correct answer is C: \(\frac{4}{5}\).

\end{tcolorbox}
\end{tabular}
\caption{A comparison of ViFT with other baseline LVLMs~(part 1).}
\end{table*}

% #######

\begin{table*}[htbp]
    \centering
    \renewcommand{\arraystretch}{1.2} 
    \setlength{\tabcolsep}{8pt} 
    \begin{tabular}{p{14cm}}
        \centering
        \includegraphics[width=0.4\textwidth]{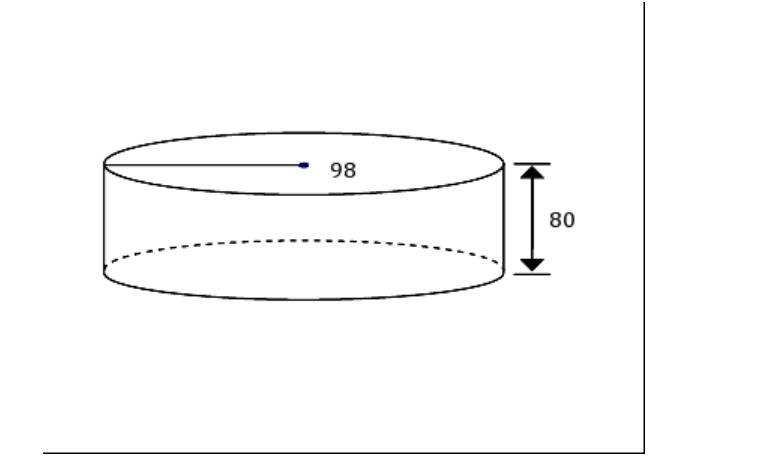}
        
        \begin{tcolorbox}[colback=gray!5, colframe=gray!40, sharp corners, width=\linewidth, boxrule=0.5mm]
        \small
        \textbf{Problem: }
        Please first conduct reasoning, and then answer the question and provide the correct answer.
        
        Find the surface area of the given cylinder. All measurements in the diagram are in mm.

        Round your answer to two decimal places.

        \textbf{Ground Truth:} 109603.88
        \end{tcolorbox}  % Reduce the vertical space between input and output
        \begin{tcolorbox}[colback=gray!5, colframe=gray!40, sharp corners, width=\linewidth, boxrule=0.5mm]
            \small
            \textbf{LLaMA-3.2-Vision-11B:}
            \\
**Step 1: Understand the problem**

The problem asks for the surface area of a cylinder with a height of 80 mm and a radius of 98 mm.

**Step 2: Identify the formula for the surface area of a cylinder**

The formula for the surface area of a cylinder is:
\[ \text{Surface Area} = 2\pi r^2 + 2\pi rh \]
where \(r\) is the radius and \(h\) is the height.

**Step 3: Plug in the values given in the problem**
Given that the radius \(r = 98\) mm and the height \(h = 80\) mm, we can substitute these values into the formula.

**Step 4: Calculate the surface area**
\[ \text{Surface Area} = 2\pi (98)^2 + 2\pi (98)(80) \]
\[ \text{Surface Area} = 19208\pi + 15680\pi = 34888\pi \approx 34888 \times 3.14159 \approx 109541.31 \, \text{mm}^2\]

**Step 5: Round the answer to two decimal places**
Rounded to two decimal places, the surface area is approximately 109541.31 mm\(^2\).

        \end{tcolorbox}

        \begin{tcolorbox}[colback=gray!5, colframe=gray!40, sharp corners, width=\linewidth, boxrule=0.5mm]
        \small
        \textbf{InternLM-XComposer-2.5-7B:}
        \\
To find the surface area of a cylinder, we need to calculate the area of the two circular bases and the lateral surface area. The formula for the area of a circle is \( A = \pi r^2 \), where \( A \) is the area and \( r \) is the radius.
First, let's find the area of one base:
\[ A_{\text{base}} = \pi r^2 = \pi (98)^2 = 9624.6 \, \text{mm}^2 \]
Since there are two bases, we double the area of one base to get the total area of both bases:
\[ A_{\text{total bases}} = 2 \times A_{\text{base}} = 2 \times 9624.6 = 19249.2 \, \text{mm}^2 \]
Next, we need to find the lateral surface area. The lateral surface area of a cylinder is given by the formula \( A = 2\pi rh \), where \( h \) is the height of the cylinder.
\[ A_{\text{lateral}} = 2\pi rh = 2\pi (98) (80) = 19200 \, \text{mm}^2 \]
Finally, we add the areas of the bases and the lateral surface area to get the total surface area of the cylinder:
\[ A_{\text{total}} = A_{\text{total bases}} + A_{\text{lateral}} = 19249.2 + 19200 = 38449.2 \, \text{mm}^2 \]

\end{tcolorbox}

\end{tabular}
\caption{A comparison of ViFT with other baseline LVLMs~(part 2).}
\label{tab:case10}
\end{table*}

\begin{table*}[htbp]
    \ContinuedFloat
    \begin{tabular}{p{14cm}}
        \centering
        
        \begin{tcolorbox}[colback=gray!5, colframe=gray!40, sharp corners, width=\linewidth, boxrule=0.5mm]
        \small
        \textbf{Qwen2-VL-7B:}
        \\
The surface area of a cylinder is given by the formula \(2\pi rh + 2\pi r^2\), where \(r\) is the radius and \(h\) is the height. Substituting the given values, we get \(2\pi \times 98 \times 80 + 2\pi \times 98^2\). Calculating this, we find the surface area to be approximately 23,567.44 mm². Therefore, the answer is 23567.44.

\end{tcolorbox}
                \begin{tcolorbox}[colback=gray!5, colframe=gray!40, sharp corners, width=\linewidth, boxrule=0.5mm]
        \small
        \textbf{LLaVA-OneVision-7B:}
        \\
The surface area of a cylinder is given by the formula: 2$\pi r(h + r)$, where r is the radius and h is the height.
Substituting the given values, we get: $2 * 3.1416 * 98 * (98 + 80) = 2 * 3.1416 * 98 * 178 = 12210.4 mm^2$
Therefore, the surface area of the cylinder is $12210.4 mm^2$.

\end{tcolorbox}
        \begin{tcolorbox}[colback=gray!5, colframe=gray!40, sharp corners, width=\linewidth, boxrule=0.5mm]
        \small
        \textbf{ViFT-7B:}
        \\
To solve this problem, we need to analyze the given geometric configuration and use properties of circles and angles.

Given:

- There is a circle with center \( A \).

- Points \( B \), \( C \), \( D \), and \( E \) lie on the circumference of the circle.

- Angle \( BAC \) and angle \( EAD \) sum up to 180 degrees.

\#\#\# Analysis:

1. **Circle Properties**: Since \( A \) is the center of the circle, any line segment from \( A \) to a point on the circumference (e.g., \( B \), \( C \), \( D \), or \( E \)) is a radius of the circle.

2. **Angle Sum Property**: The sum of opposite angles in a cyclic quadrilateral is 180 degrees. This means that if \( BAC \) and \( EAD \) are supplementary, they form a straight line when extended through the circle.

\#\#\# Steps to Solve:

To find the surface area of a cylinder, we use the formula:

\[ A = 2\pi r(h + r) \]

where:
- \( A \) is the surface area,
- \( r \) is the radius of the base,
- \( h \) is the height of the cylinder.

Given:
- Radius (\( r \)) = 98 mm,
- Height (\( h \)) = 80 mm.

Substitute these values into the formula:

\[ A = 2\pi (98)(80 + 98) \]
\[ A = 2\pi (98)(178) \]
\[ A = 2\pi \times 17444 \]
\[ A = 34888\pi \]

Now, calculate the numerical value using \(\pi \approx 3.14159\):

\[ A \approx 34888 \times 3.14159 \]
\[ A \approx 109678.472 \]

Rounding to two decimal places:

\[ A \approx 109678.47 \]

Therefore, the surface area of the cylinder is approximately 109678.47 square millimeters.

\end{tcolorbox}
\end{tabular}
\caption{A comparison of ViFT with other baseline LVLMs~(part 2).}
\end{table*}

% ########

\begin{table*}[htbp]
    \centering
    \renewcommand{\arraystretch}{1.2} 
    \setlength{\tabcolsep}{8pt} 
    \begin{tabular}{p{14cm}}
        \centering
        \includegraphics[width=0.4\textwidth]{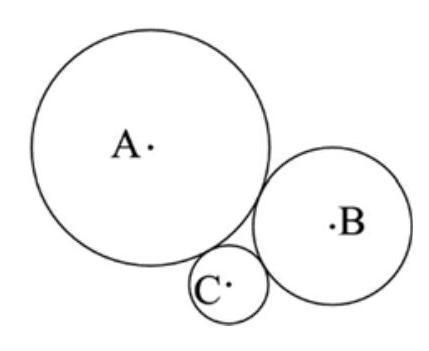}
        
        \begin{tcolorbox}[colback=gray!5, colframe=gray!40, sharp corners, width=\linewidth, boxrule=0.5mm]
        \small
        \textbf{Problem: }
        Please first conduct reasoning, and then answer the question and provide the correct answer.
        
        Three circles with centres $A, B, C$ touch each other in pairs from the outside (see diagram). Their radii are 3,2 and 1. How big is the area of the triangle $A B C$?

        \textbf{Ground Truth:} 6
        \end{tcolorbox}  % Reduce the vertical space between input and output
        \begin{tcolorbox}[colback=gray!5, colframe=gray!40, sharp corners, width=\linewidth, boxrule=0.5mm]
            \small
            \textbf{LLaMA-3.2-Vision-11B:}
            \\
To find the area of triangle $ABC$, we can use the formula for the area of a triangle given its base and height. The base of the triangle is the distance between points $A$ and $B$, and the height is the distance from point $C$ to line $AB$.

The distance between points $A$ and $B$ is equal to the sum of their radii, which is $3 + 2 = 5$. This is because the circles touch each other in pairs from the outside, so the distance between their centers is equal to the sum of their radii.

The distance from point $C$ to line $AB$ is equal to the radius of circle $C$, which is $1$. This is because point $C$ is on the circumference of circle $C$, and the radius of a circle is the distance from its center to any point on its circumference.

Now, we can use the formula for the area of a triangle:

Area = (base × height) / 2

Substituting the values we found earlier, we get:

Area = (5 × 1) / 2
= 5/2
= 2.5

Therefore, the area of triangle $ABC$ is 2.5 square units.

*Answer*: 2.5

        \end{tcolorbox}

        \begin{tcolorbox}[colback=gray!5, colframe=gray!40, sharp corners, width=\linewidth, boxrule=0.5mm]
        \small
        \textbf{InternLM-XComposer-2.5-7B:}
        \\
To find the area of triangle \(ABC\), we can use Heron's formula, which states that for a triangle with sides of length \(a\), \(b\), and \(c\), the area (\(A\)) is given by:

\[ A = \sqrt{s(s-a)(s-b)(s-c)} \]

where \(s\) is the semi-perimeter of the triangle, defined as:

\[ s = \frac{a + b + c}{2} \]

In this case, the sides of the triangle are the distances between the centers of the circles. The distance between the centers of circles with radii 3 and 2 is the sum of their radii, so it is \(3 + 2 = 5\). The distance between the centers of circles with radii 2 and 1 is \(2 + 1 = 3\), and the distance between the centers of circles with radii 3 and 1 is \(3 + 1 = 4\).

The semi-perimeter \(s\) of triangle \(ABC\) is therefore:

\[ s = \frac{5 + 3 + 4}{2} = \frac{12}{2} = 6 \]

Using Heron's formula, the area \(A\) of triangle \(ABC\) is:

\[ A = \sqrt{6(6-5)(6-3)(6-4)} = \sqrt{6 \cdot 1 \cdot 3 \cdot 2} = \sqrt{36} = 6 \]

Therefore, the area of triangle \(ABC\) is \(6\) square units.

\end{tcolorbox}

\end{tabular}
\caption{A comparison of ViFT with other baseline LVLMs~(part 3).}
\label{tab:case11}
\end{table*}

\begin{table*}[htbp]
    \ContinuedFloat
    \begin{tabular}{p{14cm}}
        \centering
        
        \begin{tcolorbox}[colback=gray!5, colframe=gray!40, sharp corners, width=\linewidth, boxrule=0.5mm]
        \small
        \textbf{Qwen2-VL-7B:}
        \\
The lengths of the sides of triangle ABC are equal to the sums of the radii of the circles that touch at those points. Therefore, the lengths of the sides of triangle ABC are 3+2=5, 2+1=3, and 3+1=4. The semi-perimeter of triangle ABC is (5+3+4)/2=6. Therefore, the area of triangle ABC is $\sqrt{(6×(6-5)×(6-3)×(6-4))}=3\sqrt{5}$. Therefore, the answer is $3\sqrt{5}$.

\end{tcolorbox}
                \begin{tcolorbox}[colback=gray!5, colframe=gray!40, sharp corners, width=\linewidth, boxrule=0.5mm]
        \small
        \textbf{LLaVA-OneVision-7B:}
        \\
The answer is 4.

\end{tcolorbox}
        \begin{tcolorbox}[colback=gray!5, colframe=gray!40, sharp corners, width=\linewidth, boxrule=0.5mm]
        \small
        \textbf{ViFT-7B:}
        \\
To find the area of triangle \(ABC\), we need to determine the lengths of its sides and then use Heron's formula or another suitable method.

Given:
- Circle \(A\) has a radius of 3 units.
- Circle \(B\) has a radius of 2 units.
- Circle \(C\) has a radius of 1 unit.

Since the circles touch each other externally, the distance between the centers of any two circles is equal to the sum of their radii.

1. **Distance between centers \(A\) and \(B\)**:
   \[
   AB = r_A + r_B = 3 + 2 = 5 \text{ units}
   \]

2. **Distance between centers \(A\) and \(C\)**:
   \[
   AC = r_A + r_C = 3 + 1 = 4 \text{ units}
   \]

3. **Distance between centers \(B\) and \(C\)**:
   \[
   BC = r_B + r_C = 2 + 1 = 3 \text{ units}
   \]

Now, we have the side lengths of triangle \(ABC\):
- \(AB = 5\) units
- \(AC = 4\) units
- \(BC = 3\) units

We can use Heron's formula to calculate the area of triangle \(ABC\). Heron's formula states that the area \(K\) of a triangle with sides of lengths \(a\), \(b\), and \(c\) is given by:
\[
K = \sqrt{s(s-a)(s-b)(s-c)}
\]
where \(s\) is the semi-perimeter of the triangle:
\[
s = \frac{a + b + c}{2}
\]

For our triangle:
\[
s = \frac{5 + 4 + 3}{2} = \frac{12}{2} = 6 \text{ units}
\]

Substituting the values into Heron's formula:
\[
K = \sqrt{6(6-5)(6-4)(6-3)} = \sqrt{6 \cdot 1 \cdot 2 \cdot 3} = \sqrt{36} = 6 \text{ square units}
\]

Therefore, the area of triangle \(ABC\) is:
\[
\boxed{6}
\]

\end{tcolorbox}
\end{tabular}
\caption{A comparison of ViFT with other baseline LVLMs~(part 3).}
\end{table*}

\begin{table*}[htbp]
    \centering
    \renewcommand{\arraystretch}{1.2} 
    \setlength{\tabcolsep}{8pt} 
    \begin{tabular}{p{14cm}}
        \centering
        \includegraphics[width=0.4\textwidth]{cases/Cases_0.jpg}
        
        \begin{tcolorbox}[colback=gray!5, colframe=gray!40, sharp corners, width=\linewidth, boxrule=0.5mm]
        \small
        \textbf{Problem: }
        Subtract all small cyan cubes. Subtract all large yellow rubber cubes. How many objects are left?

        \textbf{Ground Truth:} 8
        \end{tcolorbox}  % Reduce the vertical space between input and output
        \begin{tcolorbox}[colback=gray!5, colframe=gray!40, sharp corners, width=\linewidth, boxrule=0.5mm]
            \small
            \textbf{ViFT With Vision-Text Inputs}
            \\
The image depicts a collection of various geometric shapes, including cylinders and cubes, arranged on a flat surface. The objects vary in size, color, and material properties such as metallic or matte finishes.

Here is a detailed description of the objects present:

1. **Cylinders**:
   - There are five cylindrical objects in total.
   
   - Two large metallic cylinders: one gray and one red.
   
   - One medium-sized metallic cylinder: gold-colored.
   
   - One small metallic cylinder: blue.
   
   - One small matte cylinder: gray.

2. **Cubes**:
   - There are four cube-shaped objects.
   
   - One large metallic cube: red.
   
   - One large matte cube: gray.
   
   - One small metallic cube: cyan.
   
   - One small matte cube: cyan.

3. **Arrangement**:
   - The objects are scattered across the surface without any specific pattern.
   
   - The large metallic cylinders (gray and red) are positioned towards the back left side.
   
   - The medium-sized metallic cylinder (gold) is located near the center-left.
   
   - The small metallic cylinder (blue) is placed towards the right side.
   
   - The small matte cylinder (gray) is situated close to the small metallic cylinder (blue).
   
   - The large metallic cube (red) is positioned towards the front right.
   
   - The large matte cube (gray) is located behind the large metallic cube (red).
   
   - The two small metallic cubes (cyan) are placed in the foreground, with one slightly to the left and the other more centered.
   
   - The two small matte cubes (cyan) are also in the foreground, with one slightly to the left and the other more centered.

        \end{tcolorbox}

        \begin{tcolorbox}[colback=gray!5, colframe=gray!40, sharp corners, width=\linewidth, boxrule=0.5mm]
        \small
        \textbf{ViFT With Text-only Inputs:}
        \\
To solve this problem, we need to know the initial number of small cyan cubes and large yellow rubber cubes before any subtraction takes place. Since these values are not provided in the question, I will create a hypothetical scenario to illustrate how you would solve this problem.

Let's assume:
- There are initially 10 small cyan cubes.
- There are initially 5 large yellow rubber cubes.
- No other types of objects are mentioned.

Step 1: Subtract all small cyan cubes.
- Initial count: 10 small cyan cubes.
- After subtraction: 10 - 10 = 0 small cyan cubes.

Step 2: Subtract all large yellow rubber cubes.
- Initial count: 5 large yellow rubber cubes.
- After subtraction: 5 - 5 = 0 large yellow rubber cubes.

Step 3: Calculate the total number of objects left.
- Small cyan cubes left: 0
- Large yellow rubber cubes left: 0
- Total objects left: 0 + 0 = 0

Therefore, the final value is 0.

\end{tcolorbox}
\end{tabular}
\caption{An example of ViFT with or without ability-fused inference (part 1).}
\label{tab:case0}
\end{table*}

\begin{table*}[htbp]
    \centering
    \renewcommand{\arraystretch}{1.2} 
    \setlength{\tabcolsep}{8pt} 
    \begin{tabular}{p{14cm}}
        \centering
        \includegraphics[width=0.4\textwidth]{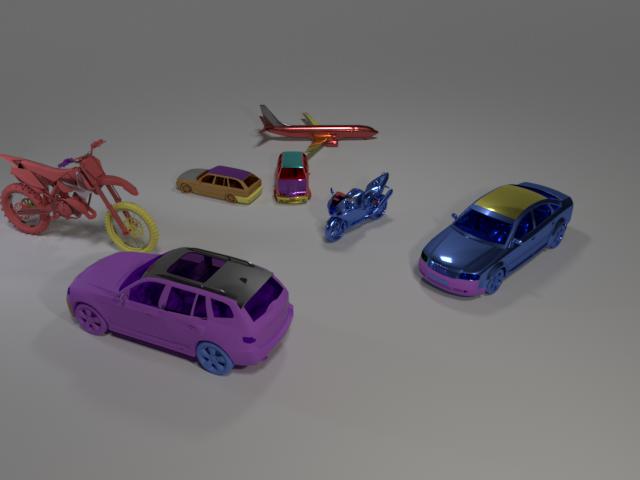}
        
        \begin{tcolorbox}[colback=gray!5, colframe=gray!40, sharp corners, width=\linewidth, boxrule=0.5mm]
        \small
        \textbf{Problem: }
        Is the number of big shiny cars in front of the red airliner greater than the number of big purple road bikes? Think first, then answer.

        \textbf{Ground Truth:} Yes
        \end{tcolorbox}  % Reduce the vertical space between input and output
        \begin{tcolorbox}[colback=gray!5, colframe=gray!40, sharp corners, width=\linewidth, boxrule=0.5mm]
            \small
            \textbf{ViFT Without Ability-Fused Inference:}
            \\
In the image, there is one big shiny car located to the right side of the red airliner. There are no big purple road bikes present in the scene. Therefore, the number of big shiny cars (1) is not greater than the number of big purple road bikes (0).

Answer: (B) No

        \end{tcolorbox}

        \begin{tcolorbox}[colback=gray!5, colframe=gray!40, sharp corners, width=\linewidth, boxrule=0.5mm]
        \small
        \textbf{ViFT With Ability-Fused Inference:}
        \\
To solve this problem, let's break it down step by step:

1. **Identify the objects**: We need to count the number of "big shiny cars" and "big purple road bikes" in relation to the red airliner.

2. **Counting the big shiny cars**:
   - There is one big shiny car (a purple SUV) in front of the red airliner.

3. **Counting the big purple road bikes**:
   - There are no big purple road bikes visible in the image.

4. **Comparison**:
   - The number of big shiny cars in front of the red airliner is 1.
   - The number of big purple road bikes is 0.

5. **Conclusion**:
   - Since 1 is greater than 0, the statement "Is the number of big shiny cars in front of the red airliner greater than the number of big purple road bikes?" is true.

Therefore, the correct answer is (A) Yes.

\end{tcolorbox}
\end{tabular}
\caption{An example of ViFT with or without ability-fused inference (part 2).}
\label{tab:case1}
\end{table*}

\begin{table*}[htbp]
    \centering
    \renewcommand{\arraystretch}{1.2} 
    \setlength{\tabcolsep}{8pt} 
    \begin{tabular}{p{14cm}}
        \centering
        \includegraphics[width=0.3\textwidth]{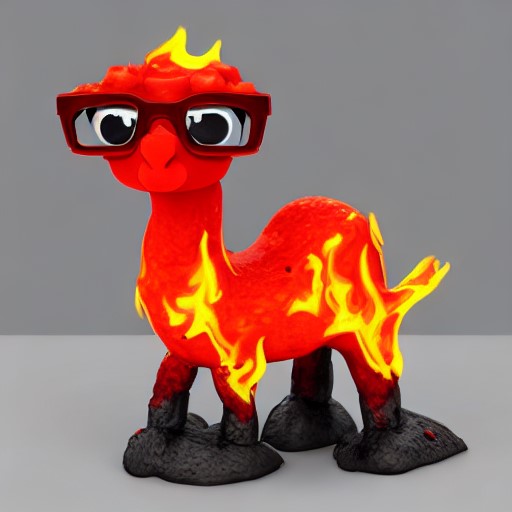}
        
        \begin{tcolorbox}[colback=gray!5, colframe=gray!40, sharp corners, width=\linewidth, boxrule=0.5mm]
        \small
        \textbf{Problem: }
        What material appears to make up the creature? Think first, then answer.

        \end{tcolorbox}  % Reduce the vertical space between input and output
        \begin{tcolorbox}[colback=gray!5, colframe=gray!40, sharp corners, width=\linewidth, boxrule=0.5mm]
            \small
            \textbf{ViFT Without Ability-Fused Inference:}
            \\ 
The image presents a 3D rendering of a creature that is a fusion of a camel and a flame. The creature stands on four legs, each ending in black hooves. Its body is predominantly red, with yellow flames dancing across its surface, giving it an appearance reminiscent of a living fire. 

The creature's head is adorned with a pair of glasses, adding a touch of whimsy to its fiery form. The background is a simple gray gradient, which serves to highlight the vibrant colors of the creature. The creature is positioned centrally in the image, drawing the viewer's attention immediately to its unique blend of elements. 

There are no discernible texts or other objects in the image. The relative position of the creature to the background suggests it is standing still, with no indication of movement or action. The image does not provide any additional context or narrative, leaving the interpretation open to the viewer.

        \end{tcolorbox}
        
        \begin{tcolorbox}[colback=gray!5, colframe=gray!40, sharp corners, width=\linewidth, boxrule=0.5mm]
        \small
        \textbf{ViFT With Ability-Fused Inference:}
        \\
To determine what material appears to make up the creature in the image, let's analyze its characteristics:

1. **Color and Texture**: The creature has a vibrant red body with yellow and orange flames emanating from it. This suggests that the material might be something that glows or emits light, possibly indicating heat or energy.

2. **Shape and Structure**: The creature resembles an animal, specifically a camel, but with exaggerated features like large eyes and a flame-like appearance. Its legs are sturdy and end in black hooves.

3. **Material Properties**:
   - **Red Body**: The red color could indicate a material like clay, plastic, or even a painted surface.
   - **Flames**: The presence of flames suggests that the material could be designed to mimic fire or glow-in-the-dark properties.
   - **Sturdy Legs**: The legs appear solid and robust, which might suggest a material like metal or hard plastic.

4. **Contextual Clues**: Given the fantastical nature of the creature (a flaming camel), it is likely a representation rather than a real-life object. This implies that the material could be anything used for artistic purposes, such as clay, plastic, or even digital rendering.

5. **Lighting and Shadows**: The way the light interacts with the creature suggests a smooth, possibly glossy surface, which could be indicative of materials like plastic or polished stone.

Considering these points, the most plausible material would be a combination of clay or plastic for the body, with additional elements like paint or lighting effects to create the flame effect. The sturdy legs could be made of a different material, perhaps metal or hard plastic, to provide stability.

In conclusion, while the exact material cannot be definitively determined without further information, the creature seems to be crafted from a combination of clay/plastic for the body and possibly metal or hard plastic for the legs, with added visual effects to simulate flames.

\end{tcolorbox}
\end{tabular}
\caption{An example of ViFT with or without ability-fused inference (part 3).}
\label{tab:case2}
\end{table*}

\begin{table*}[htbp]
    \centering
    \renewcommand{\arraystretch}{1.2} 
    \setlength{\tabcolsep}{8pt} 
    \begin{tabular}{p{14cm}}
        \centering
        \includegraphics[width=0.5\textwidth]{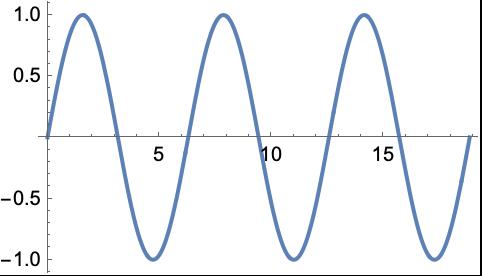}
        
        \begin{tcolorbox}[colback=gray!5, colframe=gray!40, sharp corners, width=\linewidth, boxrule=0.5mm]
        \small
        \textbf{Problem: }
        What is this function most likely be? Think first, then answer.

        \textbf{Ground Truth:} B. a trigonometric function
        \end{tcolorbox}  % Reduce the vertical space between input and output
        \begin{tcolorbox}[colback=gray!5, colframe=gray!40, sharp corners, width=\linewidth, boxrule=0.5mm]
            \small
            \textbf{$\alpha=0.8$:}
            \\
However, without the actual graph or additional context, I cannot provide a specific identification of the function. The options given (polynomial, trigonometric, exponential, or logarithmic) suggest different types of functions that could fit the description. If you have more details about the graph or the function itself, please provide them so I can give a more accurate identification.

        \end{tcolorbox}

        \begin{tcolorbox}[colback=gray!5, colframe=gray!40, sharp corners, width=\linewidth, boxrule=0.5mm]
        \small
        \textbf{$\alpha=0.9$:}
        \\
To determine which type of function the given graph represents, let's analyze its characteristics:

1. **Shape**: The graph shows a smooth, periodic wave-like pattern with peaks and troughs that repeat at regular intervals.

2. **Amplitude**: The peaks and troughs appear to reach up to 1 unit above and below the x-axis.

3. **Periodicity**: The pattern repeats every 5 units along the x-axis.

Given these observations, we can infer the following:

- **Polynomial Functions**: These typically have a more complex shape with multiple turning points (peaks and troughs), but they do not exhibit the repetitive nature seen in the graph.

- **Trigonometric Functions**: These functions, such as sine or cosine, are known for their periodic behavior, which matches the observed pattern in the graph.

- **Exponential Functions**: These functions grow or decay exponentially and do not form a repeating pattern like the one shown.

- **Logarithmic Functions**: These functions also do not exhibit the periodicity seen in the graph.

Based on the analysis, the most likely function represented by the graph is a trigonometric function.

Therefore, the correct answer is (B) a trigonometric function.

\end{tcolorbox}

        \begin{tcolorbox}[colback=gray!5, colframe=gray!40, sharp corners, width=\linewidth, boxrule=0.5mm]
        \small
        \textbf{$\alpha=1.0$:}
        \\
To determine what type of function the given graph represents, let's analyze its characteristics:

1. **Shape**: The graph shows a smooth, periodic wave-like pattern with peaks and troughs that repeat at regular intervals.

2. **Amplitude**: The peaks and troughs appear to reach approximately ±1 on the y-axis.

3. **Periodicity**: The pattern repeats every 5 units along the x-axis, indicating a period of 5.

Given these observations, we can infer that the function is likely a sine or cosine function, which are fundamental trigonometric functions known for their wave-like behavior.

- **Polynomial Functions**: These typically have a smooth curve but do not exhibit the repetitive nature seen in the graph.

- **Exponential Functions**: These usually show rapid growth or decay and do not form a repeating pattern.

- **Logarithmic Functions**: These also do not exhibit the periodicity observed in the graph.

Therefore, based on the shape and periodic nature of the graph, the most likely function is a trigonometric function.

Answer: (B) a trigonometric function

\end{tcolorbox}

\end{tabular}
\caption{An example of ViFT with different $\alpha$ for ability fusion.}
\label{tab:case4}
\end{table*}

\begin{table*}[htbp]
    \ContinuedFloat
    \begin{tabular}{p{14cm}}
        \centering

                \begin{tcolorbox}[colback=gray!5, colframe=gray!40, sharp corners, width=\linewidth, boxrule=0.5mm]
        \small
        \textbf{$\alpha=1.1$:}
        \\
The given graph depicts a periodic oscillation with a sinusoidal pattern. The function appears to be a sine or cosine wave, which are fundamental trigonometric functions. These functions are characterized by their smooth, repetitive curves that rise and fall in a regular manner.

To determine the specific type of trigonometric function, we can analyze the shape and properties of the curve:

1. **Periodicity**: The graph shows a repeating pattern every 5 units along the x-axis. This indicates that the period of the function is 5 units.

2. **Amplitude**: The maximum value of the function is 1, and it reaches this peak twice within each period (once at the top and once at the bottom).

3. **Shape**: The curve starts at zero, rises to its maximum positive value, then decreases back to zero, continues to its minimum negative value, and finally returns to zero before starting another cycle.

Given these characteristics, the function is best described as a trigonometric function, specifically a sine or cosine function. Since the graph starts at the origin and reaches its first peak at \( x = 5 \), it suggests that the function could be a sine function shifted horizontally.

Therefore, the most likely function represented in the graph is a trigonometric function.

Answer: (B)

\end{tcolorbox}
        \begin{tcolorbox}[colback=gray!5, colframe=gray!40, sharp corners, width=\linewidth, boxrule=0.5mm]
        \small
        \textbf{$\alpha=1.2$:}
        \\
The given graph depicts a periodic oscillation with a sinusoidal pattern. The function appears to be a sine or cosine wave, which are fundamental trigonometric functions. These functions are characterized by their smooth, repetitive curves that rise and fall in a regular manner.

To determine the specific type of trigonometric function, we can analyze the shape and properties of the curve:

1. **Periodicity**: The graph shows a repeating pattern every 5 units along the x-axis. This indicates that the period of the function is 5 units.

2. **Amplitude**: The maximum value of the function is 1, and it reaches this peak twice within each period (once at the top and once at the bottom).

3. **Shape**: The curve starts at zero, rises to its maximum positive value, then decreases back to zero, continues to its minimum negative value, and finally returns to zero before starting another cycle.

Given these characteristics, the function is best described as a trigonometric function, specifically a sine or cosine function. Since the graph starts at the origin and reaches its first peak at \( x = 5 \), it suggests that the function could be a sine function shifted horizontally.

Therefore, the most likely function represented in the graph is a trigonometric function.

Answer: (B)

\end{tcolorbox}
\end{tabular}
\caption{An example of ViFT with different $\alpha$ for ability fusion.}
\end{table*}

\begin{table*}[htbp]
    \centering
    \renewcommand{\arraystretch}{1.2} 
    \setlength{\tabcolsep}{8pt} 
    \begin{tabular}{p{14cm}}
        \centering
        \includegraphics[width=0.6\textwidth]{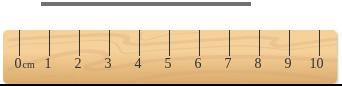}
        
        \begin{tcolorbox}[colback=gray!5, colframe=gray!40, sharp corners, width=\linewidth, boxrule=0.5mm]
        \small
        \textbf{Problem: }
        Move the ruler to measure the length of the line to the nearest centimeter. How long is the line? Think first, then answer.

        \textbf{Ground Truth:} 7
        \end{tcolorbox}  % Reduce the vertical space between input and output
        \begin{tcolorbox}[colback=gray!5, colframe=gray!40, sharp corners, width=\linewidth, boxrule=0.5mm]
            \small
            \textbf{$\beta=0$:}
            \\
The image shows a ruler with measurements in centimeters, ranging from 0 to 10 cm. There is a black line above the ruler that spans from approximately 2 cm to just past the 6 cm mark. Therefore, the length of the line is about 5 centimeters.

        \end{tcolorbox}

        \begin{tcolorbox}[colback=gray!5, colframe=gray!40, sharp corners, width=\linewidth, boxrule=0.5mm]
        \small
        \textbf{$\beta=0.1$:}
        \\
To determine the length of the line using the ruler provided, we need to follow these steps:

1. **Identify the Scale**: The ruler has markings from 0 cm to 10 cm.

2. **Align the Line with the Ruler**: Place one end of the line at the 0 cm mark on the ruler.

3. **Measure the Length**: Extend the line along the ruler until it reaches the other end.

From the image, the line starts at the 0 cm mark and extends to just past the 7 cm mark but not reaching the 8 cm mark. Therefore, the line measures approximately 7 centimeters.

**Answer:** The line is about 7 centimeters long.

\end{tcolorbox}

        \begin{tcolorbox}[colback=gray!5, colframe=gray!40, sharp corners, width=\linewidth, boxrule=0.5mm]
        \small
        \textbf{$\beta=0.2$:}
        \\
To determine the length of the line using the ruler provided, we need to follow these steps:

1. **Identify the Scale**: The ruler has markings from 0 cm to 10 cm.

2. **Align the Line**: Place the ruler such that one end of the line aligns with the 0 cm mark on the ruler.

3. **Measure the Length**: Extend the ruler along the line until it reaches the other end of the line.

Given the hint to provide an integer answer, let's assume the line extends to just past the 7 cm mark but not reaching 8 cm. Therefore, the length of the line would be approximately 7 centimeters.

**Answer:** 7

\end{tcolorbox}

        \begin{tcolorbox}[colback=gray!5, colframe=gray!40, sharp corners, width=\linewidth, boxrule=0.5mm]
        \small
        \textbf{$\beta=0.3$:}
        \\
To determine the length of the line using the ruler provided, we need to follow these steps:

1. **Identify the Scale**: The ruler has markings from 0 cm to 10 cm.

2. **Align the Line**: Place the ruler such that one end of the line aligns with the 0 cm mark on the ruler.

3. **Measure the Length**: Extend the ruler along the line until it reaches the other end of the line.

Given the hint to provide an integer answer, let's assume the line extends to just past the 7 cm mark but not reaching 8 cm. Therefore, the length of the line would be approximately 7 centimeters.

**Answer:** 7

\end{tcolorbox}

        \begin{tcolorbox}[colback=gray!5, colframe=gray!40, sharp corners, width=\linewidth, boxrule=0.5mm]
        \small
        \textbf{$\beta=0.4$:}
        \\
!!!!!!!!!!!!!!!!!!!!!!!!!!!!!!!!!!!!!!!!!!!!!!!!!!!!!!!!!!!!!!!!!!!!!!!!!!!!!!!!!!!!!!!!!!!!!!!!!!!!!!!!!!!!!!!!!!!!!!!!

\end{tcolorbox}
\end{tabular}
\caption{An example of ViFT with different $\beta$ for ability fusion.}
\label{tab:case3}
\end{table*}

\end{document}